\pdfoutput=1
\documentclass[11pt]{article}

\usepackage[final]{acl}
\usepackage{fdsymbol}

\usepackage{times}
\usepackage{color} % comment
\usepackage{latexsym}
\usepackage{makecell}
\usepackage{tablefootnote}
\usepackage[toc,page]{appendix}
\usepackage{pifont}% http://ctan.org/pkg/pifont
\newcommand{\cmark}{\ding{51}}%
\newcommand{\xmark}{\ding{55}}%

\usepackage{enumitem}
\usepackage[T1]{fontenc}
\usepackage[utf8]{inputenc}

\usepackage{microtype}

\usepackage{hyperref}
\usepackage{booktabs}
\usepackage{graphicx}
\usepackage{makecell}
\graphicspath{{./imgs/}}
\usepackage{footmisc}

\usepackage{floatrow}
\usepackage{microtype}

\usepackage{caption}
\usepackage{subcaption}
\usepackage{array, makecell} %

\usepackage{pifont}% http://ctan.org/pkg/pifont
\usepackage{graphics}

\usepackage{tabularx,colortbl}

\usepackage{tikz}
\usepackage{collcell}

\usepackage{etoolbox}

\newcolumntype{?}{!{\vrule width 1.5pt}}

\newtoggle{inTableHeader}% Track if still in header of table
\toggletrue{inTableHeader}% Set initial value
\newcommand*{\StartTableHeader}{\global\toggletrue{inTableHeader}}%
\let\OldTabular\tabular%
\let\OldEndTabular\endtabular%
\renewenvironment{tabular}{\StartTableHeader\OldTabular}{\OldEndTabular\StartTableHeader}%

\newcommand*{\MinNumber}{-1.0}%
\newcommand*{\MidNumber}{0.0} %
\newcommand*{\MaxNumber}{1.0}%

\newcommand{\ApplyGradient}[1]{%
  \iftoggle{inTableHeader}{#1}{
    \ifdim #1 pt > \MidNumber pt
        \pgfmathsetmacro{\PercentColor}{max(min(100.0*(#1 - \MidNumber)/(\MaxNumber-\MidNumber),100.0),0.00)} %
        \hspace{-0.33em}\colorbox{yellow!\PercentColor!blue}{#1}
    \else
        \pgfmathsetmacro{\PercentColor}{max(min(100.0*(\MidNumber - #1)/(\MidNumber-\MinNumber),100.0),0.00)} %
        \hspace{-0.33em}\colorbox{blue!\PercentColor!blue}{#1}
    \fi
  }}
\newcolumntype{R}{>{\collectcell\ApplyGradient}c<{\endcollectcell}}

\usepackage{amsmath}
\usepackage{amsfonts,bm}
\usepackage{xspace}

\newcommand{\red}[1]{\textcolor{red}{#1}}
\newcommand{\blue}[1]{\textcolor{blue}{#1}}

\newcommand{\Ni}{({\em i})~}
\newcommand{\Nii}{({\em ii})~}

\definecolor{mypink3}{cmyk}{0, 0.7808, 0.4429, 0.1412}

\makeatletter   
\newcommand{\sveryshortarrow}[1][3pt]{\mathrel{%
    \vcenter{\hbox{\rule[-.5\fontdimen8\scriptfont3]
               {\scriptratio\dimexpr#1\relax}{\fontdimen8\scriptfont3}}}%
   \mkern-4mu\hbox{\let\f@size\sf@size\usefont{U}{lasy}{m}{n}\symbol{41}}}}
\makeatother

\def\eqref#1{equation~\ref{#1}}
\def\1{\bm{1}}

\def\m1{{\bm{1}}}

\DeclareMathAlphabet{\mathsfit}{\encodingdefault}{\sfdefault}{m}{sl}
\SetMathAlphabet{\mathsfit}{bold}{\encodingdefault}{\sfdefault}{bx}{n}

\usepackage[nameinlink]{cleveref}
\crefformat{section}{\S#2#1#3} % see manual of cleveref, section 8.2.1
\crefname{algorithm}{Alg.}{Algs.}
\crefformat{subsection}{\S#2#1#3}
\Crefname{equation}{Eq.}{Eqs.}
\Crefname{figure}{Fig.}{Figs.}

\usepackage[colorinlistoftodos,prependcaption,textsize=tiny]{todonotes}

\usepackage{soul}

\usepackage{float}
\definecolor{azure}{rgb}{0.0, 0.5, 1.0}
\definecolor{darkbrown}{rgb}{0.4, 0.26, 0.13}
\usepackage{multirow}% http://ctan.org/pkg/multirow
\usepackage{hhline}% http://ctan.org/pkg/hhline

\newcommand{\model}{{ChartInstruct}}

\title{{\model}: Instruction Tuning for Chart Comprehension and Reasoning}

\author{
Ahmed Masry$^{\clubsuit}$\thanks{\ \ Equal contribution.}, \ Mehrad Shahmohammadi$^{\clubsuit}$\footnotemark[1], \\ \bf \ Md Rizwan Parvez$^{\diamondsuit}$, \ Enamul Hoque$^{\clubsuit}$, \ Shafiq Joty$^{\vardiamondsuit\spadesuit}$\\
$^\clubsuit$York University, Canada, 
$^\diamondsuit$Qatar Computing Research Institute (QCRI) \\
$^\vardiamondsuit$Salesforce Research, $^\spadesuit$Nanyang Technological University, Singapore
\\
\{msm97, enamulh\}@yorku.ca, mparvez@hbku.edu.qa \\
sjoty@salesforce.com, ahmed.elmasry24653@gmail.com
}

\begin{document}
\maketitle

\begin{abstract} 

Charts provide visual representations of data and are widely used for analyzing information, addressing queries, and conveying insights to others. Various chart-related downstream tasks have emerged recently, such as question-answering and summarization. A common strategy to solve these tasks is to fine-tune various models originally trained on vision tasks language. However, such task-specific models are not capable of solving a wide range of chart-related tasks, constraining their real-world applicability. 
To overcome these challenges, we introduce \textit{\model}:  a novel chart-specific vision-language Instruction-following dataset comprising 
191K instructions generated with 71K
charts. We then present two distinct systems for instruction tuning on such datasets: (1) an end-to-end model that connects a vision encoder for chart understanding with a LLM; and (2) a pipeline model that employs a two-step approach to extract chart data tables and input them into the LLM.  In experiments on four downstream tasks, we first show the effectiveness of our model--achieving a new set of state-of-the-art results. Further evaluation shows that our instruction-tuning approach supports a wide array of real-world chart comprehension and reasoning scenarios, thereby expanding the scope and applicability of our models to new kinds of tasks. 
\end{abstract}

\section{Introduction}

\begin{figure*}[t!]
     \centering
        % \vspace{mm}
        \includegraphics[width=.98\textwidth]{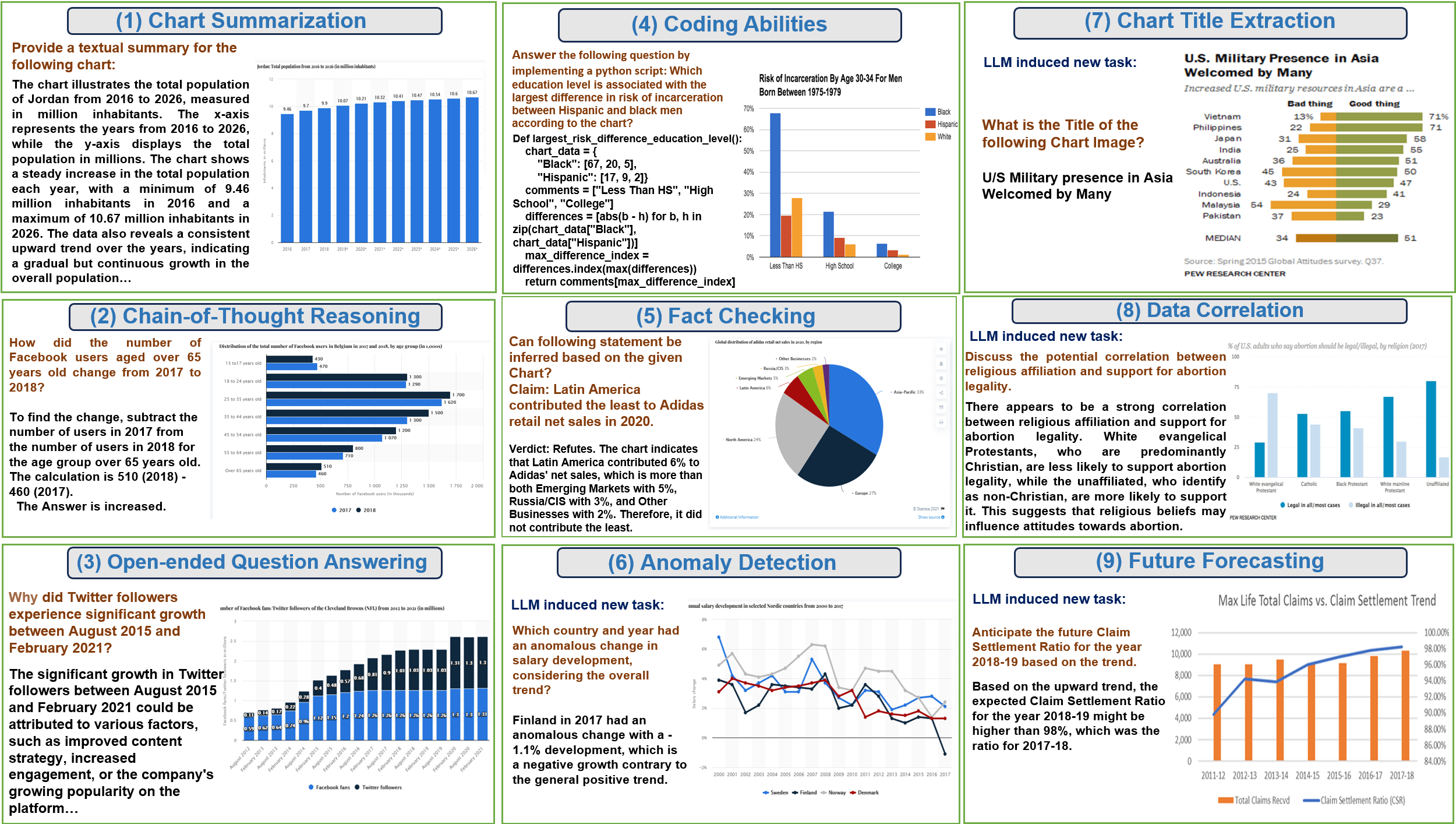}
        %\vspace{-3mm}
         \caption{\small{ Examples of 
         different chart-related tasks from our generated instruction dataset. Examples 1-5 are generated based on predefined tasks similar to previously developed downstream tasks like chart summarization, chart question answering, and fact-checking, while examples 6-9 introduce new types of tasks distilled by LLMs. 
         }
         %\vspace{-3mm}
          }
    \label{fig:ITune}
\end{figure*}

Information visualizations, such as bar charts and line charts, play a pivotal role in data analysis, offering critical insights and aiding in informed decision-making \cite{hoque2022chartSurvey}. However, discerning key patterns and trends from these visualizations and addressing complex queries can pose significant challenges. Recent research has introduced various tasks to assist users in chart analysis, including chart question answering \citep{masry-etal-2022-chartqa, open-CQA}, summarizing insights from visualizations \citep{obeid-hoque-2020-chart, chart-to-text-acl}, reasoning over chart images, fact-checking \citep{akhtar2023reading}, and automated visual data storytelling \citep{shi2020calliope}.

Early work attempts to tackle these tasks by fine-tuning models originally trained on language and vision tasks \citep{Raffel2020t5, masry-etal-2022-chartqa, lee2022pix2struct, kantharaj-etal-2022-chart}. However, such models may not be optimal for chart-specific tasks as they overlook explicit modeling of chart structures such as relationships between chart elements like bars, legends, and axes. Recent models such as UniChart \citep{masry2023unichart}, Chart-T5 \citep{zhou-etal-2023-enhanced}, and MatCha \citep{liu2022matcha} are specifically designed for charts by considering visual and mathematical reasoning over chart elements and values. However, they often consider charts from a limited range of sources and focus on a narrow set of tasks, constraining their real-world applicability. Indeed, for real-world widespread adoption, we cannot presume how and on what tasks these models will be used. 

A promising solution to this challenge is \emph{instruction tuning}, as demonstrated by language models like InstructGPT \cite{instructgpt}, FLAN-T5 \cite{chung2022scaling},  Alpaca \citep{Alpaca}, \citep{Vicuna} and LLaMA-chat \cite{touvron2023llama}. 
They show that training LLMs on instruction-following datasets significantly enhances their alignment with user intent across various tasks, including  tasks that are unseen during training. Recent advances in vision-language tasks \citep{li2023llava, dai2023instructblip} have adopted similar methodologies, fine-tuning vision-language models (VLMs) with visual instructions to better match user intentions and improve efficacy. However, to our knowledge, instruction tuning for chart comprehension and reasoning remains  underexplored. Existing methods \citep{liu2023mmc, han2023chartllama}, which are concurrent to our work, lack variety and scope in instruction-tuning tasks, limiting their effectiveness in real-world chart understanding scenarios. 

In this paper, we introduce Chart Instruction Tuning (\textit{\model}), to pave the way towards building general-purpose chart comprehension and reasoning assistant based on VLMs. To this end, we have developed a new chart instruction-tuning dataset featuring real-world charts collected from 157 online platforms, covering wide and diverse visual styles. Leveraging advanced LLMs such as GPT-3.5\citep{Chatgpt}
, GPT-4 \citep{openai2023gpt4},
and Gemini \citep{geminiteam2023gemini}), we  generate \textbf{191K} instructions covering a broad array of tasks reflecting real-world applications (Figure~\ref{fig:ITune}).

As charts are unique and pose challenges distinct from general multi-modal data, a structured approach is crucial for enabling VLMs to effectively leverage the instruction dataset, optimizing their performance in chart analysis tasks.  We introduce two innovative VLM designs in this regard. The first system modifies the LLaVA architecture \citep{li2023llava}, substituting its CLIP vision encoder with the UniChart \citep{masry2023unichart} vision encoder pre-trained specifically on chart images.  
For language modeling in this design, we experiment with two different models: Llama2 (7B) -- a decoder-only model \citep{touvron2023llama} and Flan-T5-XL (3B) -- an encoder-decoder model \citep{chung2022scaling}.
Our second design adopts a two-step pipeline approach:  first extract the underlying data table from the chart image, then provide it as input to the LLM.  This range of models provides a spectrum of efficient solutions, making our systems adaptable to various real-world scenarios and computational demands. 

Our comprehensive evaluation across \textbf{four} benchmarks: ChartQA \citep{MSMmasry2022chartqa}, Chart2Text \citep{charttotext}, OpenCQA \citep{open-CQA}, ChartFC \citep{akhtar2023reading}
demonstrates our models' state-of-the-art performance in chart understanding and reasoning tasks.  
Human evaluation further suggests the effectiveness of our instruction-tuning approach in supporting a wide array of real-world chart comprehension and reasoning scenarios, broadening its adaptability to numerous new tasks.

Our main contributions include:  (i) A new instruction-following corpus with real-world charts and a wide range of tasks by utilizing LLMs, (ii) two distinct systems 
specifically tailored for chart understanding tasks; (iii) extensive 
evaluations that demonstrate the state-of-the-art performance of \model \space across existing 
chart-related benchmark tasks while also expanding its applicability to new tasks.
We have made our code and chart corpus publicly available at \href{https://github.com/vis-nlp/ChartInstruct}{https://github.com/vis-nlp/ChartInstruct}.

\begin{figure*}[t!]
     \centering
        % \vspace{mm}
        \includegraphics[width=.98\textwidth]{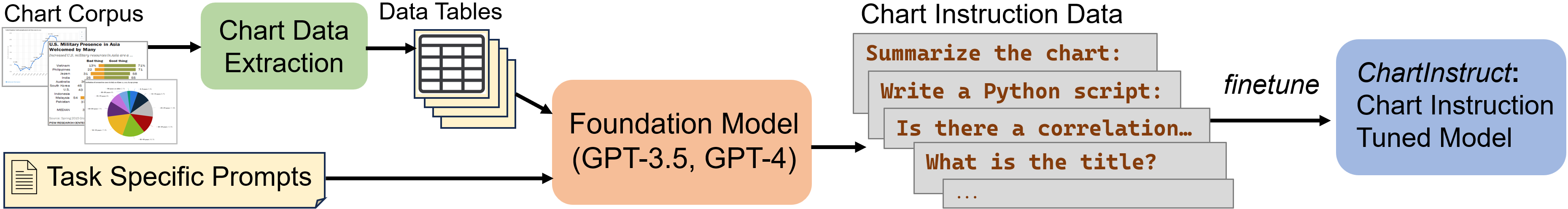}
         \caption{\small{
         Instruction tuning process for chart collection. For the WebChart Corpus, the chart data is extracted automatically using Gemini Vision Pro. For distilling new tasks we use GPT-4, for other task generation we either use GPT 3.5 or GPT 4. 
         }
         \vspace{-2mm}
         }
    \label{fig:instruct-tuning-process}
\end{figure*}
\section{Related Work}

\subsection{Chart Modeling}
Chart understanding methods fall into two main categories: those directly fine-tuned from language or vision-language models~\cite{MSMmasry2022chartqa, ahmed-workshop-2021, lee2022pix2struct} and those specifically crafted for chart-specific tasks \cite{masry2023unichart, liu2022matcha, zhou-etal-2023-enhanced}. Models in the former category often exhibit limited performance due to a lack of chart-specific pretraining. In contrast, models in the latter group are pretrained on primitive chart-specific tasks like question answering and summarization which constrains their applicability across diverse real-world chart scenarios. Recent works \cite{wang2023domino, do2023llms, huang2023lvlms} have also employed LLMs such as GPT-3, Llama \cite{touvron2023llama}, and GPT-4 for chart-related tasks. They utilize pipelines that extract data values from chart images using specialized models like UniChart \cite{masry2023unichart} and Deplot \cite{liu2022deplot}, which are then used by the LLMs in different downstream tasks like question answering and summarization. However, these methods either rely on proprietary models \cite{GPT-3, openai2023gpt4} or public models without chart-specific training, limiting their effectiveness and generalization.

\subsection{Visual Instruction Tuning}

Instruction tuning in LLMs has shown benefits in aligning models with human intent and enhancing task generalization \cite{chung2022scaling, NEURIPS2022_b1efde53, wang2023selfinstruct, Alpaca, Vicuna}. These techniques have also been extended to the vision-language space~\cite{li2023llava, zhu2023minigpt, ye2023mplugowl, li2023blip, dai2023instructblip}. However, visual instruction tuning approaches in the chart domain are rare. Although a few studies have implemented multimodal instruction tuning \cite{liu2023mmc, han2023chartllama}, they depend on CLIP \cite{clipvisual} for vision encoding, designed for natural images rather than charts. Moreover, their training often relies on synthetic charts or a narrow selection of real-world charts, focusing on a narrow set of instruction-tuning tasks. In contrast, our work presents two systems explicitly designed for the chart domain, trained on a diverse array of chart images and covering a wide range of real-world chart applications.
 
\subsection{Chart Domain Downstream Tasks}

Interest in chart-related tasks is rising, focusing on understanding and generating information from charts. Chart question answering (CQA) addresses queries about charts with some datasets \cite{plotqa} and \cite{masry-etal-2022-chartqa} focusing on visual and arithmetic reasoning, while others focus on open-ended explanatory question answering (OpenCQA) \cite{open-CQA}. Additionally, Chart-to-Text involves creating natural language summaries from charts \cite{chart-to-text-acl}, and Chart-to-Table focuses on converting charts into data tables \cite{Choi2019VisualizingFT, masry2023unichart}. Automated Fact-Checking (AFC) for images, including charts, aims to check claims against data \cite{akhtar2023reading, akhtar2023chartcheck}. In this paper, we evaluate our models on these different downstream tasks and also generate new kind of chart reasoning tasks through the instruction data generation process.

\begin{table*}[t]
 \setlength\extrarowheight{1pt}
 \centering

 \scalebox{0.67}{\begin{tabular}{lcccccc|cc}
  
  \toprule
  
  \multirow{1}{*}{Dataset
  } & \multirow{1}{*}{CoT Reasoning} & \multicolumn{1}{c}{Chart Summarization} & 
   \multicolumn{1}{c}{Fact Checking} & \multicolumn{1}{c}{Open-ended QA}  & \multicolumn{1}{c}{Coding Abilities}& \multicolumn{1}{c}{Novel Tasks} & \multicolumn{1}{c}{\#Unique Charts}\\

  \midrule

  Statista   & 4,363 & 4,159 & 4,188 
  &4,906   & 2,348 & - & 9992\\ 

  PlotQA   & 4,159 & 3977 & 4,333 
  & 12,105  & 2,306 & - & 8199\\

  OECD/OWID   &4290 & 3999   
  & 4,080 & 13,213  & 2,994 & - & 10949\\

  WebCrawled   & 14,459 & 41741 & 11,574 
  & 12,246  & 11,924 & 23,410 & 41742\\
  \midrule
  Total  & 27,271(14.3\%) & 53,876(28.24\%) & 24,175(12.67\%) & 42,470(22.26\%)  & 19,572(10.26\%) & 23,410(12.27\%) & 70,882
  \\ \bottomrule

 \end{tabular}}
 \vspace{-3mm}
 \caption{\small {The number of generated examples for each
 tasks based on data samples of the mentioned dataset. Some of the charts are used in multiple tasks. On the last Column, we show the number of distinct charts used for instruction generation samples.
 }
 \vspace{-3mm}
 }
 \label{tab:instructiondata}
\end{table*}

\section{Chart Instruction Data Generation}

We build an instruction tuning dataset for enhancing VLMs' capabilities 
in tackling diverse understainding and generation tasks related to chart analysis.  In this section, we describe the chart corpus collection followed by the instruction tuning data generation process. Figure~\ref{fig:instruct-tuning-process} provides an overview of the %chart 
instruction tuning process. 
%}

\subsection{Chart Corpora Collection}
Our goal is to build a diverse chart dataset using real-world data to enhance our model's generalizability. 
In order to build that we collect chart images from two main sources: existing public datasets and web-crawled charts. We chose UniChart \cite{masry2023unichart} from existing datasets, as it provides one of the largest and most diverse chart pretraining corpora, containing 611K charts with 
%associated 
metadata like data tables, titles, and captions (refer to \citet{masry2023unichart} for details). However, these charts come from a few specific online sources such as Pew \cite{pew}, Statista \cite{statista}, OECD \cite{oecd}, and OWID \cite{owid}, limiting the variety of visual styles and data domains covered.

To address this limitation, we contribute with a new corpus, WebCharts, which contains \textbf{41K}  diverse chart images. 
We started with a seed list of web domains containing charts \cite{hoque2019searching} and then use the top image search results from these domains using queries such as "chart images," "graphs," and "visual data".
We then develop a binary VIT classifier \cite{vit} to distinguish chart images from non-chart images in our search results, followed by manual removal of any remaining non-chart images to refine the dataset. However, these charts lack the underlying data tables which are critical for instruction generation on various chart data analysis tasks. 
Therefore, we automatically extract the data tables and chart titles using Gemini Pro Vision \cite{geminiteam2023gemini}. The choice of Gemini was influenced by the cost and the unlimited API rate features. More details about the web charts collection process are provided in Appendix \ref{appendix:webchart}.

\subsection{Instruction Data Generation}
To enhance LLMs' performance in chart-related tasks via instruction tuning,  we develop our chart instruction Dataset. This dataset has 190,774 instructions corresponding to 70,882 charts, covering various aspects of chart comprehension and reasoning (see examples in Figure \ref{fig:ITune}). 
Below, we describe the process of generating instruction data.

\textbf{\Ni Tasks Selection:} To cover diverse aspects of chart reasoning and comprehension, we identify a set of tasks that are similar to some existing downstream tasks such as chart summarization and question answering (QA) but also included other tasks such as code generation and Chain of thought reasoning. Additionally, we prompted LLMs to propose novel tasks to enrich the dataset. Below we briefly explain these tasks.

\noindent \textbf{$\bullet$ Summarization and QA} The summarization task aims to generate a chart caption that captures the key insights such as trends and patterns from a given chart~\cite{kantharaj-etal-2022-chart}. We also include Open-ended QA~\cite{kantharaj-etal-2022-opencqa} in which the model generates an explanatory answer to the given question about a chart.

\noindent \textbf{$\bullet$ Fact Checking} task~\cite{akhtar2023reading} is included to improve our model's ability to reduce errors and interpret chart data accurately. It takes a claim as input and then generates the verdict (`refute' or `accept') along with an explanation. 

\noindent \textbf{$\bullet$ Chain-of-thought (CoT) Reasoning} aims to enhance the model's ability to perform complex mathematical and visual reasoning. 
%involving multiple steps. 
However, open-source models (e.g.,  \citet{touvron2023llama} and \citet{jiang2023mistral}) often incur errors in mathematical computations. To address this, we devised two types of questions: (1) Variable Dependent questions, which use tools to compute statistics, inspired by ToolFormer \cite{schick2023toolformer}, and (2) Variable Independent Questions, focusing on retrieval, comparison, and basic math analysis.

\noindent \textbf{$\bullet$ Code Generation} task is included to generate executable Python scripts to answer user queries, drawing inspiration from the success of this approach demonstrated by PAL \citep{gao2023pal}.

\noindent \textbf{$\bullet$ Novel Tasks} play a crucial role in enhancing the diversity of the instruction set. We tasked an LLM to propose different possible chart-related tasks. To prevent overlap with existing tasks like chart summarization and QA, we instructed the LLM not to replicate these tasks. The generated instructions involve tasks that may require new forms of reasoning and analysis (e.g., future value predictions).

\textbf{\Nii Prompt Design:} To create the instructions for different tasks, we first design a set of prompt templates, where each template contains: (1) \textbf{task description}, (2) the \textbf{input} chart data table, along with metadata such as the chart title, (3) output \textbf{constraints} (if any), (4) \textbf{output format}. An example prompt is shown in Figure~\ref{fig:Prompt_sample}.

\textbf{\Nii Input and Output Generation:} After designing prompts, we generate instructions, creating input-output instances for each prompt template. 
We concatenate each chart's underlying data table and title with one of the prompts designed for the expected task by utilizing OpenAI's GPT3.5 Turbo and GPT4 (see Table \ref{tab:datamodel}). The choice between these APIs for different tasks is dictated by the task complexity, with GPT-4 being employed for complex reasoning tasks, and GPT3.5 for tasks with moderate complexity. Moreover, in order to reduce the generation cost, and increase the variety of generated samples in each call, in our prompts, we ask the LLMs to come up with multiple samples for each chart. 
The input prompt for each 
%specific 
task is provided through Tables \ref{tab:FCsample}-\ref{tab:Novelsample} in  Appendix \ref{appendix:generatedsamples}.

\begin{figure}[t!]
    \centering
    \includegraphics[width=0.9\textwidth]{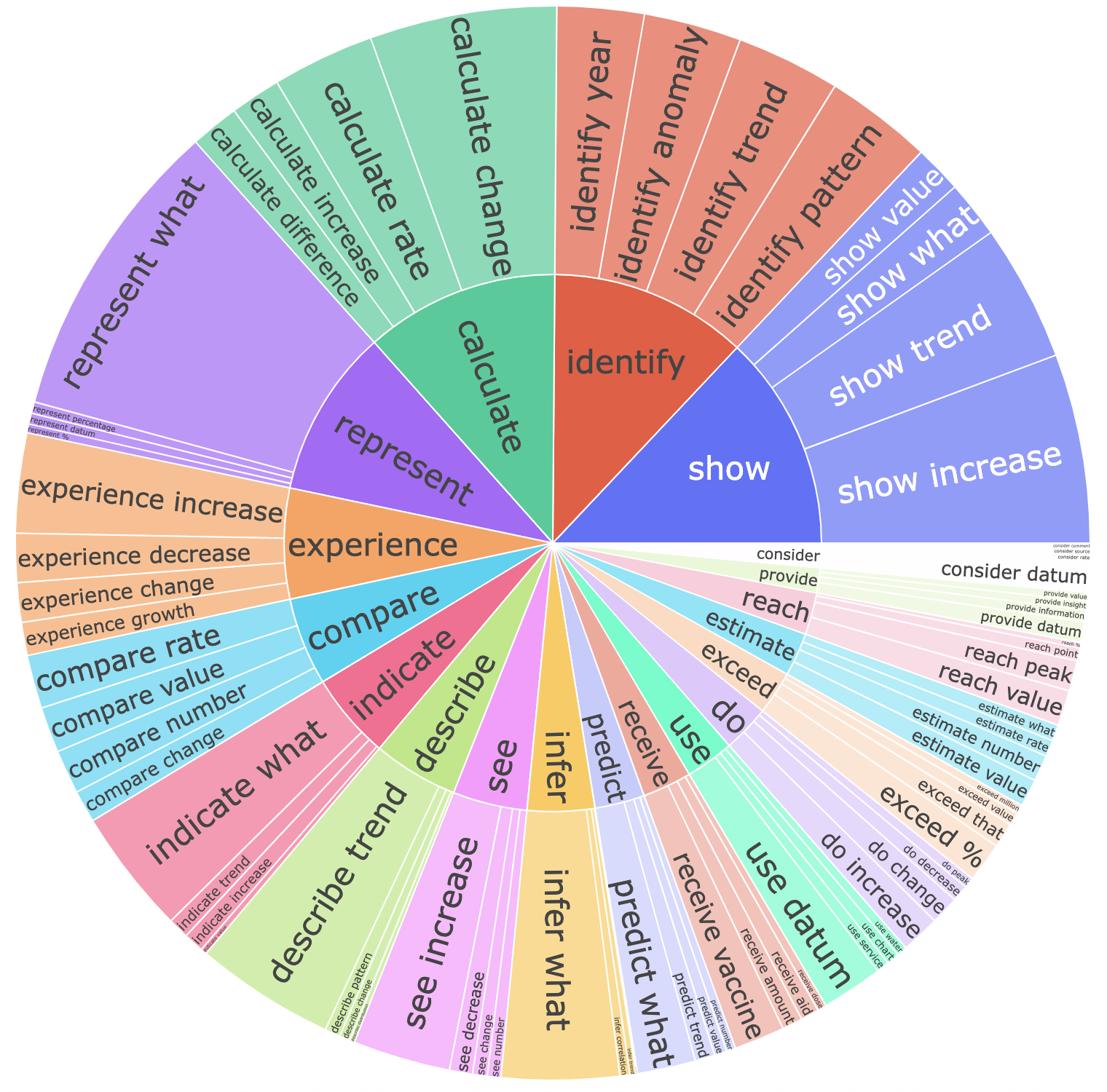}
    \caption{Top 20 most common root verb (inner circle) and corresponding four object verb pairs for all the generated instructions of our dataset.
    }
    \label{fig:subburty}
    \vspace{-2mm}
\end{figure}

\subsection{Dataset Analysis}
We present key statistics and analyze the diversity and quality of the instruction dataset.

\textbf{Statistics:} Our Chart corpus (WebCharts) is highly diverse, encompassing a variety of bar and line charts, pie and donut charts, and even unconventional chart types not prevalent in existing chart corpora (see Figure ~\ref{fig:chart_types}). The generated instructions set is dominated by Chart summarization and open-ended QA to improve the chart comprehension ability but also augmented with reasoning tasks and creative new tasks generated by LLM (see Table \ref{tab:instructiondata}). We placed particular emphasis on WebCharts dataset due to its diversity, constituting 67.5\% (157,190 samples) of our dataset.

\textbf{Diversity:}  To investigate the diversity of generated instructions, %and the range of tasks covered, 
we employed the Berkeley Neural Parser \cite{Berkeley} to identify the verb closest to the root along with its first direct noun object in each instruction.
The analysis reveals a broad spectrum of comprehension and reasoning tasks expressed in the instructions (Figure~\ref{fig:subburty}). 
We further analyzed the newly proposed tasks generated by GPT-4 by clustering the instructions using the K-means algorithm. From Table~\ref{tab:NovelCounts} in Appendix \ref{appendix:wccorpus}, we observe that Pattern and Outlier Detection is the most common type of task, followed by various statistical analyses. Notably, the dataset includes interesting tasks not typically captured by existing downstream tasks, such as identifying correlations, predicting values and trends, and distribution analysis. Overall, it suggests that the generated instructions set is indeed diverse and creative. We also visualize diversity in the length of the instructions' inputs, and instance outputs in Figure \ref{fig:inplen} and \ref{fig:outlen}. 

\textbf{Quality:} We asked an expert annotator to evaluate the quality of the generated data on a random set of 100 instructions. We find that in general, the instructions describe a valid task (87\%)
%\mehrad{87} 
and the input matches the task description (86\%) among generated instructions. In 61\% and 8\% cases, outputs for the generated inputs were fully and partially correct respectively. 
We list a number of correct and incorrect examples in Figure \ref{fig:few-generated-samples}. We note that even when the outputs may be incorrect (e.g., contain factual errors), the corresponding task instructions can provide informative training signal as found by others (e.g., \citet{honovich2022unnatural}).

\section{Modeling}

In this section, we describe our two architectures for chart instruction tuning.

\subsection{End-to-End System}

\begin{figure}[t!]
     \centering
        % \vspace{mm}
        \includegraphics[width=1\textwidth]{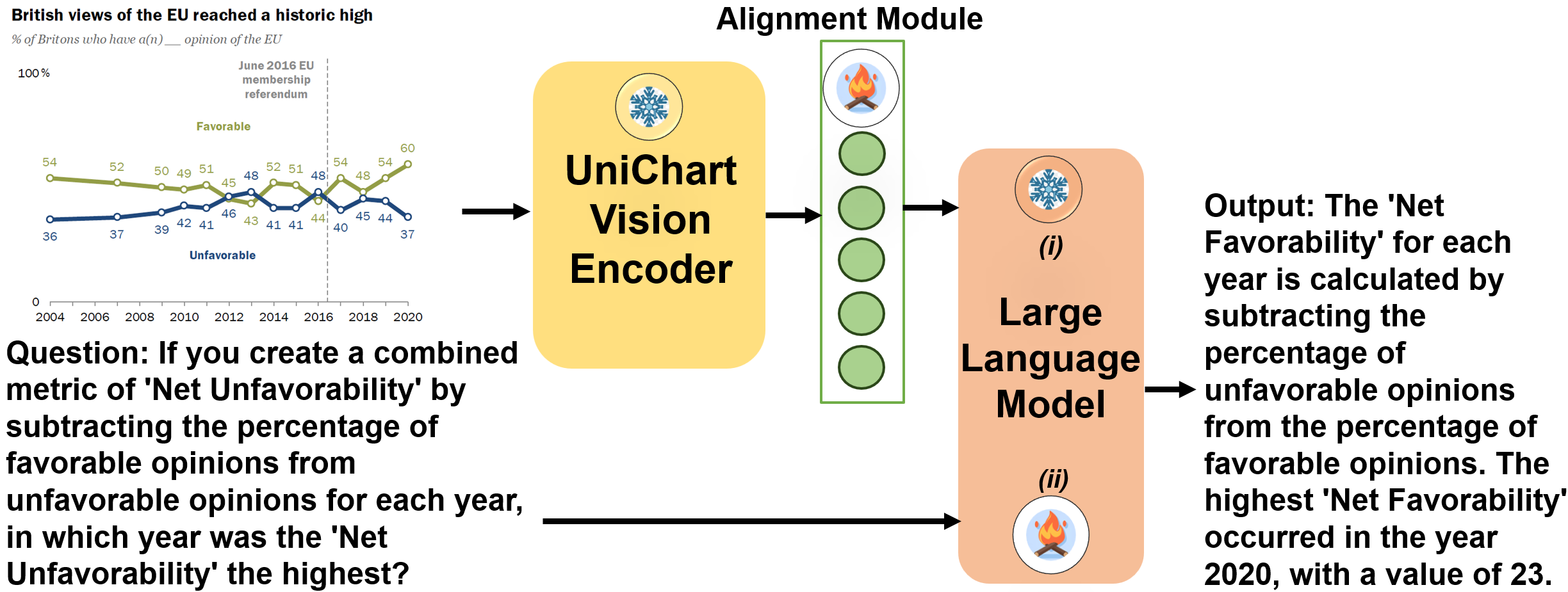}
        \vspace{-3mm}
         \caption{\small{
         The architecture for our end-to-end system models: the LLM is frozen in our (i) pre-training step, while it updates its parameters in the (ii) instruction-tuning step. We either use Flan-T5-XL or Llama2 as LLM for this architecture. We show our pipeline system architecture in \ref{fig:new_pipeline} in \ref{appendix:model}} 
         }\label{fig:end-to-end-both}
         \vspace{-4mm}
\end{figure}

Our end-to-end system utilizes the LLaVA \cite{liu2023visual} architecture, which incorporates CLIP \cite{clipvisual} for visual encoding, an LLM for language generation, and an adapter module for transforming the encoded visual features to the LLM's input embedding space (Figure 
\ref{fig:end-to-end-both}).
LLaVA  was originally designed for natural image understanding. We made the following modifications to adapt it for chart understanding.
First, we substituted the CLIP vision encoder with the UniChart vision encoder \cite{masry2023unichart}, which is pretrained and optimized for chart image understanding. For the LLM, we investigate two model types: the  decoder-only architecture of Llama2 \cite{han2023chartllama} and  encoder-decoder structure of Flan-T5 \cite{chung2022scaling}. In the Llama2 setup, projected visual features are injected directly into the language decoder, whereas in the Flan-T5 model, these features, along with the instructions, are first processed by the language encoder before the decoder 
generates 
a text. We experimented with both the 7B variant of Llama2 and the 3B variant of Flan-T5 to provide a range of model sizes suitable for different applications.

In this end-to-end design, before fine-tuning the model on instructional data, we first {fine-tune} only the  adaptor module keeping the vision encoder and  LLM 
frozen. This critical \emph{alignment stage} is necessary to align the visual features from the UniChart vision encoder with the input embedding space of the LLM, and enables the LLM to accurately interpret chart images. We focus on two specific tasks for this phase: generating data tables from charts and summarizing chart contents. After alignment, we finetune the model on instruction tuning data, keeping the vision encoder frozen while training the weights of both the adaptor and LLM.

\subsection{Pipeline System}
In this approach, a data table generation module first converts the chart image into a textual data table representation.
This generated data table is then combined with the input instruction and fed into an LLM. We utilize UniChart  \cite{masry2023unichart}, which has been shown to be able to generate high-quality data tables from chart images, ensuring the textual representation closely mirrors the original chart's information. For the  LLM, we conduct experiments with both Llama2, and Flan-T5 models similar to our end-to-end approach. Unlike the end-to-end system, this setup skips the alignment step since the visual features are not directly fed into the LLM. Hence, we directly finetune the models on the instruction %tuning
data (see Figure~\ref{fig:new_pipeline}).

\begin{table*}[t]
 \setlength\extrarowheight{2pt}
 \centering
\resizebox{\linewidth}{!}{
 \scalebox{0.60}{\begin{tabular}{l|c|ccccccc}
  
  \toprule
  
  \multirow{3}{*}{} & \multirow{3}{*}{} & \multicolumn{3}{c}{ChartQA} & 
  OpenCQA & \multicolumn{2}{c}{Chart-to-Text}  & ChartFC \\ 
  \multirow{3}{*}{} & \multirow{3}{*}{} & \multicolumn{3}{c}{\small ($RA$)} & \small ($BLEU$) & \multicolumn{2}{c}{\small ($BLEU$)}  & \small ($Accuracy$) \\
  \cmidrule(lr){3-5} \cmidrule(lr){7-8}
  
   Model & \#Params & aug. & human & avg. & OpenCQA & Pew & Statista  & ChartFC \\

  \midrule
  VisionTaPas \cite{masry-etal-2022-chartqa} & - & 61.44 & 29.60 & 45.52 & -- & - & -  & - \\
  
  T5 \cite{masry-etal-2022-chartqa} & 222M & 56.96 & 25.12 & 41.04 
  & 9.28 & 10.49 & 35.29 & - \\
  
  VL-T5 \cite{masry-etal-2022-chartqa} & - & 56.88 & 26.24 & 41.56 
  & 14.73 & - & - & - \\

  ChartBERT \cite{akhtar2023reading} & - & - & - & -
  & - & - & - & \underline{63.8} \\
  
  Pix2Struct \cite{lee2022pix2struct} & 282M & 81.6 & 30.5 & 56.0 
  & - & 10.3 & 38.0 & - \\
  
  Matcha\cite{liu2022matcha}  & 282M & \underline{90.2} & 38.2 & 64.2 
  & - & 12.2 & \underline{39.4} & - \\ 

  UniChart \cite{masry2023unichart}  & 201M & 88.56 & \underline{43.92} & \underline{66.24} & \underline{14.88} 
  & \underline{12.48} & 38.21 & - \\

  \midrule

  \multicolumn{9}{l}{\textbf{End-to-End System}} \\

  ChartInstruct-Flan-T5-XL  & 3B & 85.04 & 43.36 & 64.2 
  & \textbf{16.71} & 12.92 & 42.42 & 70.27 \\
                                   
  ChartInstruct-Llama  & 7B & 87.76 & 45.52 & 66.64
  & 15.59 & \textbf{13.83} & \textbf{43.53} & 69.57 \\ 
  
  \midrule
  \multicolumn{9}{l}{\textbf{Pipeline System}} \\

  ChartInstruct-Flan-T5-XL  & 3B & \textbf{93.84} & \textbf{50.16} & \textbf{72.00} 
  & 14.81 & 9.93 & 40.08  & \textbf{72.65} \\

  ChartInstruct-Llama  & 7B & 82.40 & 40.64 & 61.52
  & 14.78 & 12.81 & 39.39 & 64.99 \\ 
  \bottomrule
 \end{tabular}}
 }
 \vspace{-3mm}
 \caption{\small {Evaluation results on four public benchmarks: ChartQA, Chart-to-Text, OpenCQA, and ChartFC. All the results are calculated after finetuning \model. 
 }
 \vspace{-3mm}
 }
 \label{tab:results}
\end{table*}
\section{Experiments and Results}

We evaluate the usefulness of \model\ and demonstrate that our models built upon \model\ achieve excellence in chart understanding and generation tasks.  In addition to the existing downstream tasks, they also posit superior capabilities in new tasks. Below, we first discuss the setups, then experiments on downstream benchmarks and new chart tasks, and finally present an error analysis and challenges. Our evaluation complements the trivial quantitative approach based on automated metrics with detailed human evaluation on both seen and new tasks in multiple aspects, reflecting the true understanding of effectiveness of \model\ . To ensure the reproducability of our research, we present the hyperparameters of our instruction tuning and downstream tasks experiments in Table \ref{tab:hyperparameters}. All experiments were conduced on a 4-A100 GPUs (80GB) machine.

\subsection{Experimental Setup}

\textbf{Downstream Tasks}:
%\label{sec:downstream-tasks}
To assess the generalizability of our models across a spectrum of practical chart applications,  
we evaluate them on \textbf{four} established downstream tasks in the literature: (i) ChartQA \cite{MSMmasry2022chartqa} -- a factoid chart question answering dataset, (ii) OpenCQA \cite{open-CQA} -- an open-ended chart question answering dataset, (iii) Chart2Text \cite{chart-to-text-acl} -- a chart captioning dataset collected from two sources: Statista \cite{statista} and Pew Research Center \cite{pew},  and (iv) ChartFC \cite{akhtar2023reading} --  a chart fact checking dataset. 
Furthermore, we conduct a human evaluation to explore their adaptability in real-world scenarios beyond these benchmarks.

\noindent{\bf Baselines: } \label{subsec:baselines}
We compare \model\ against seven baselines: (1) \emph{T5} \cite{Raffel2020t5}, a unified seq2seq Transformer model; (2) \emph{VL-T5} \cite{vlt5}, a T5-based model for Vision-Language (VL) tasks; (3)~\emph{VisionTapas} \cite{masry-etal-2022-chartqa}, an extension of TaPas \cite{tapas} for chart question answering; (4)~\emph{ChartBERT} \cite{akhtar2023chartcheck}, a BERT-based model utilizing textual and visual information of charts for fact verification; (5)~\emph{Pix2Struct} \cite{lee2022pix2struct}, a pretrained image-to-text model; (6) \emph{MatCha} \cite{liu2022matcha}, an adaptation of Pix2Struct for charts pretrained on math reasoning; and (7) \emph{UniChart} ~\cite{masry2023unichart}, achieving SoTA on Chart-to-Text, ChartQA, and OpenCQA.

\noindent{\bf Evaluation Metrics} We use Relaxed Accuracy (RA) for ChartQA (following \citet{plotqa}), Accuracy for ChartFC~\cite{akhtar2023chartcheck} and BLEU for text-generation tasks (Chart-to-Text and OpenCQA) \cite{post-2018-call}. However, BLEU focuses mainly on n-gram matching, overlooking factors like informativeness and factual correctness \cite{goyal2022news}. To address this, we conduct human evaluations to compare these aspects (see Section \Cref{sec:human_eval}).

\subsection{Results and Findings}

We present the experimental results on downstream tasks in \Cref{tab:results} and compare with existing baselines. 
\model\ models (both end-to-end and pipeline) 
outperforms previous state-of-the-art models, UniChart on all ChartQA and Chart-to-Text datasets. In particular, the Flan-T5-XL version excels on the ChartQA including the challenging human-written question set~\cite{masry-etal-2022-chartqa}, which suggests that the model learned more complex mathematical and visual reasoning through the relevant instruction tuning tasks such as CoT reasoning, and coding abilities.  \model\ also achieved a higher BLUE score compared to UniChart on OpenCQA benchmark, which demonstrates our model's capability to generate explanatory answers for questions about charts. Finally, \model\ surpasses ChartBERT by a wide margin (8.85\%) on the recently released fact-checking task. Overall, these results establish \model\ as the SoTA model for chart comprehension and reasoning tasks.

Our observations reveal that the end-to-end system for ChartInstruct-LLama generally surpasses the corresponding LLama pipeline system across all benchmarks. This performance disparity is likely due to the fact that the data table alone does not capture all the nuanced information present in the charts, thus becoming a limiting factor in the pipeline system's effectiveness. Similarly, the end-to-end system of ChartInstruct-Flan-T5-XL performs better than the pipeline system on both OpenCQA and Chart-to-Text benchmarks. One notable exception is the reasoning-intensive tasks like ChartFC and ChartQA on which the pipeline Flan-T5-XL system exhibits better performance. Furthermore, we notice that both ChartInstruct-Flan-T5-XL and ChartInstruct-Llama achieve comparable performance, even tho the former has 4B fewer parameters. This efficiency makes ChartInstruct-Flan-T5-XL more suitable for real-world applications with computational constraints.

To further assess the impact of our different instruction tuning tasks on our model's performance, we conducted ablation studies on the ChartQA dataset using our best performing model, ChartInstruct-Flan-T5-XL (Pipeline System). Our ablation studies reveal that  excluding tasks like Chart Summarization or Open-ended Question Answering results in a minor decline in performance (\Cref{tab:ablations}). This performance dip becomes significantly pronounced upon the removal of the reasoning tasks (CoT and Coding), emphasizing their pivotal role in enhancing the model's reasoning capabilities.  More details about the experiments can be found in \Cref{app:ablations}.

\subsection{Human Evaluation on Chart Tasks}
\label{sec:human_eval}

Reference-based evaluation metrics like BLEU-score may not align with human-perceived text quality attributes \cite{liu2023geval, smith-etal-2016-climbing}. To ensure accurate evaluation of our approach, we conducted a human experiment, assessing the generated responses from UniChart and our \model-Llama model across three metrics: (a) Informativeness, (b) Relevance, and (c) Factual Correctness.

For the study, we chose 150 samples that are unseen by both UniChart and \model-Llama. Half of them are randomly from the ChartQA test set, while the other half is from a small set of web-crawled charts not used in the instruction generation pipeline. These samples contain queries from Open-ended QA, Chart Summarization, and novel instruction samples that involve a diverse set of tasks for evaluation. In terms of task distribution, 75 (50\%) of the study samples belonged to novel tasks, while the other half comprised Chart-to-Text and OpenCQA tasks (40 and 35 samples). We use UniChart and \model-Llama to generate responses for these samples. We asked 2 different annotators to rate the sample's responses based on the mentioned factors from 1-5, having 100 samples in common to measure their agreement level\footnote{ We found Cohen's Kappa of 0.447 as the agreement level.} toward the responses.  We presented the responses randomly to prevent any biases toward models. 

From Table \ref{tab:evaluation-table-human}, we observe that  \model-Llama significantly outperforms UniChart across all three measures of human evaluation, especially in relevance (4.06 vs. 2.74). Upon manual examination, we observed that UniChart often provides a general summary of the chart without addressing specific task instructions (sometimes repeating the same tokens), particularly evident in novel and OpenCQA task samples. In contrast, \model-Llama consistently offers relevant answers for these cases (see an example in Figure \ref{fig:modelcomp}). Overall, these findings affirm that our instruction-tuning approach enhances the model's ability to adhere to task instructions, thereby expanding its capacity to address a wide array of new real-world chart-related scenarios beyond the capabilities of the state-of-the-art pre-trained model for the chart domain.

\begin{table}[t!]
\vspace{-0.3em}
\setlength{\tabcolsep}{3pt}
    \centering
    \small
    \scalebox{0.75}{\begin{tabular}{lccc}
    \toprule 
    %\textbf{Summary}
    & \textbf{Informativeness} & \textbf{Relevance} & \textbf{Factual} \\ \midrule
    UniChart\cite{masry2023unichart} &    3.2 &    2.74 & 2.756  \\
    \model-Llama &    3.848 &    4.06 &   3.664
    \\

    \midrule
    $p-value$ & $7.43 \times 10^{-4}$ & $4.42 \times 10^{-5}$ & $1.31 \times 10^{-8}$  \\

    \bottomrule
    \end{tabular}
    }
    \vspace{-2mm}
    \captionsetup{width=0.97\linewidth}
    \caption{\small Human evaluation results for comparing between the outputs of UniChart and \model-Llama. The first two rows show the average of samples across each metric. The last row shows the p-values resulted from performing Mann-Whitney U Tests.
    }
    \label{tab:evaluation-table-human}
\end{table}

\subsection{Error Analysis and Challenges}

We  reviewed our model's results across various samples to highlight the challenging aspects encountered.

\noindent{\bf Value Estimation and Comparison}
Charts with crowded or minimal details pose challenges in pairing visual elements (e.g., bars) with their associated values, estimating data values, and making comparisons based on visual attributes (compare based on height). For instance, errors occurred in Q1 and Q2 of Figure \ref{fig:WrongSamples}, where the correct value associated with specified items was not identified.

\noindent{\bf Factual Errors}
Although our models have shown improved text generation quality and better utilization of available information, they still produce statements unsupported by the chart or factually incorrect. In Q3 of Figure \ref{fig:WrongSamples}, 
the model produces coherent text but also introduces factual errors.

\noindent{\bf Numerical Reasoning}
Despite advancements, LLMs sometimes struggle with dependable mathematical operations \cite{gao2022pal, liu2022matcha}. While achieving state-of-the-art performance in ChartQA \cite{masry-etal-2022-chartqa} and attempting to teach the model to use external tools, LLMs still exhibit inconsistencies in calculations. Q4 in Figure \ref{fig:WrongSamples} illustrates the unreliability of LLMs in some numerical reasoning tasks.

\section{Conclusion}
We present \model, an automatically generated dataset of chart-related instructions and two instruction systems designed for a broad range of chart-related tasks.  To the best of our knowledge, this is the first instruction tuned dataset that not only includes pre-defined tasks but also many new types of tasks automatically distilled by LLMs.  Our model sets the state-of-the-art performance on four different downstream tasks on various automatic measures while the human evaluation further confirms the effectiveness of our approach on many new kinds of tasks. We believe that our models and instruction-tuning dataset will be valuable resources for future research and encourage further exploration into the unique problem domain of chart understanding and reasoning.

\section*{Limitations}

First, while our research covers key tasks such as Chart Summarization, Chart Question-Answering, Open-ended Chart Question-Answering, and Chart Fact Checking, it does not cover other tasks, e.g., Chart-to-table. Second, while our manual inspection of instruction-tuning dataset suggests that the novel tasks distilled by LLM are generally valid and answerable, occasionally the outputs are incorrect which may influence the instruction-tuning process. Third, although our instruction tuning approach significantly enhances the model's ability to follow instructions compared to its counterpart, it does not entirely prevent the model from deviating from instructions. Fourth, despite the state-of-the-art performance on the numerical reasoning task, ChartQA, our model still struggle with complex numerical questions. Finally, the model may produce factually incorrect statements in the text generation tasks.

\section*{Ethics Statement}

During the dataset collection process, we were mindful of several ethical considerations. The first three sources of our chart corpus ( Statista\footnote{\href{https://www.statista.com/getting-started/publishing-statista-content-terms-of-use-and-publication-rights}{https://www.statista.com/getting-started/publishing-statista-content-terms-of-use-and-publication-rights}}, OWID\footnote{\href{https://ourworldindata.org/faqs\#can-i-use-or-reproduce-your-data}{https://ourworldindata.org/faqs\#can-i-use-or-reproduce-your-data}}, OECD\footnote{\href{https://www.oecd.org/termsandconditions/}{https://www.oecd.org/termsandconditions/}}) grant publication rights for academic use of their content. Moreover, the PlotQA dataset \cite{plotqa} is a publicly available dataset published under MIT license\footnote{\href{https://github.com/NiteshMethani/PlotQA}{https://github.com/NiteshMethani/PlotQA}}. We plan to release the chart images collected from these resources along with their metadata. For the WebCharts corpus, we plan to release only the URLs from which the chart images were collected following relevant large-scale vision-language datasets (e.g., LAION\footnote{\href{https://laion.ai/}{https://laion.ai/}}). Furthermore, we release our models for only academic research purposes. 

In our commitment to exclude harmful content from our chart images, we employed Google search for sourcing the chart images, leveraging its strict policies against harmful content\footnote{\href{https://blog.google/products/search/when-and-why-we-remove-content-google-search-results/}{https://blog.google/products/search/when-and-why-we-remove-content-google-search-results/}}. Moreover, all sourced chart images underwent an initial automatic filtering process using a chart classifier, followed by a manual review phase. Additionally, the WebCharts images were processed through the Gemini API, which is designed to block unsafe content\footnote{\href{https://ai.google.dev/docs/safety\_setting\_gemini}{https://ai.google.dev/docs/safety\_setting\_gemini}}, thereby providing an additional layer of assurance regarding the appropriateness of the content included in our dataset.

Given the generative nature of our models, there is a potential risk of misuse where users can generate harmful or factually incorrect outputs, potentially leading to the spread of misinformation. We urge users to exercise responsibility and caution, restricting their use of our models to academic and research purposes only. 

The human evaluation was performed by the authors and their collaborators who were involved with this research. As the focus of the research was solely on assessing models' capabilities, effectiveness, and limitations in several chart understanding tasks, the human evaluation performed by the authors does not add any ethical issues or unwanted biases. We present our instructions to the human evaluators as well as a sample in Figures \ref{fig:evalRule} and \ref{fig:evalenv}, respectively. Moreover, there were no paid participants involved in the study.
Finally, no information has been used that can directly relate to the identification of any person while evaluating the responses from the models.

\bibliographystyle{acl_natbib}
\bibliography{chart2text}
\newpage

\appendix

\begin{appendices}

\section{Chart Instruction Data Generation}
\label{appendix:graph}
\subsection{Chart Corpora Collection}
\label{appendix:webchart}

\textbf{WebCharts Collection}: For collecting charts, we conducted image queries on Google and collected chart images from 157 unique source domains. The list of keywords we use are: "chart images," "charts ","graphs," "visual data", and "data visualization". In each query, we included the web domain  (e.g., "site: nytimes.com charts") and retrieved the top image search results. We developed a binary VIT classifier \cite{vit} to distinguish chart images from non-chart images in our search results. For training, we manually labeled 1,200 images and split them into 8:1:1 ratios for train, validation, and test sets. The classifier achieved 91\% accuracy on the test set. Using this classifier, we filtered out non-chart images from the WebCharts dataset, followed by manual removal of any remaining non-chart images to finalize the dataset.  However, these charts lacked the underlying data tables necessary for instruction generation. Therefore, we automatically extracted the data tables and chart titles using Gemini Pro Vision \cite{geminiteam2023gemini}.

\subsection{LLMs used for Instruction Generation}

Table \ref{tab:datamodel} shows the models used to generate the data for different tasks.

\begin{table}[b]
 \setlength\extrarowheight{1pt}
 \centering

 \scalebox{0.9}{\begin{tabular}{lcc}
  
  \toprule

 \multirow{1}{*}{Downstream Task/Model} & \multirow{1}{*}{GPT-3.5
 } & \multirow{1}{*}{GPT-4
 } \\
  \midrule

   Chart Summarization & \cmark & \xmark \\
   Open-ended QA & \cmark & \xmark \\
 Fact Checking & \cmark & \xmark  \\
  Chain-of-Thought Reasoning & \xmark & \cmark \\
   Code Generation & \cmark & \cmark \\
 Novel tasks & \xmark & \cmark \\
  \midrule
 \end{tabular}}
 \vspace{-3mm}
 \caption{\small {Models used to generate the data for each different task. Choices are based on task complexity and costs. 
 }
 \vspace{-3mm}
 }
 \label{tab:datamodel}
\end{table}

\subsection{Input Prompts for Instruction Generation}
\label{appendix:prompts}

\begin{figure}[t!]
    \centering
    \scalebox{0.95}{\includegraphics[width=1\textwidth]{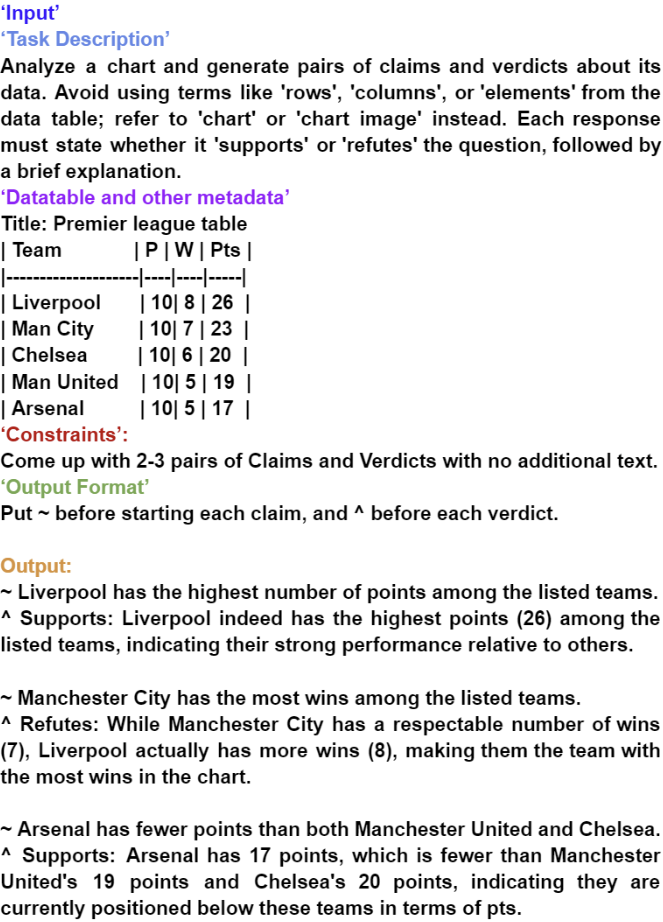}}
    \caption{    
    An example prompt to LLM and the corresponding output for a fact-checking task. The input consists of the task description, chart data, any output constraints, and output format. 
    }
    \label{fig:Prompt_sample}
\end{figure}

Figure \ref{fig:Prompt_sample} shows an example prompt to LLM and the corresponding output for a fact-checking task.

\begin{figure}[t!]
    \centering
    \scalebox{0.98}{\includegraphics[width=1\textwidth]{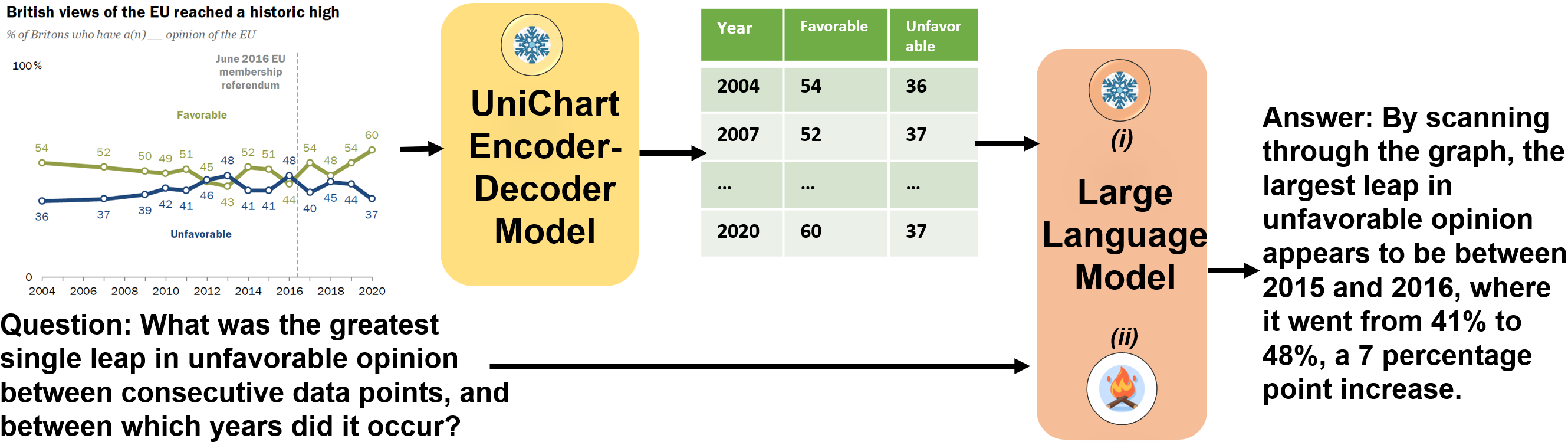}}
    \caption{    The architecture for our pipeline models:  the encoder is frozen, while LLM updates its parameters. 
    }
    \label{fig:new_pipeline}
\end{figure}

\begin{table*}[ht]
 \setlength\extrarowheight{2pt}
 \centering
 \caption{
  Training details for our instruction tuning and downstream tasks finetuning experiments.
 }
 \scalebox{0.7}{\begin{tabular}{l|cccc|cccc}
  
  % \cline{1-6}
    \toprule
    & \multicolumn{4}{c}{\textbf{End-to-End System}} & \multicolumn{4}{c}{\textbf{Pipeline System}} \\
    \midrule
   \textbf{Experiment} & \textbf{\# Epochs} & \textbf{Learning Rate} & \textbf{Batch Size} & \textbf{Hours} & \textbf{\# Epochs} & \textbf{Learning Rate} & \textbf{Batch Size} & \textbf{Hours} \\ \midrule

    & \multicolumn{8}{c}{\textbf{Alignment}}  \\ \midrule
   Flan-T5-XL & 4 & 2e-3 & 128 & ~7 & - & - & - & - \\
   Llama 2 & 3  & 2e-3 & 64 & ~24 & - & - & - & - \\ \midrule
   
    & \multicolumn{8}{c}{\textbf{Instruction Tuning}}  \\ \midrule
   Flan-T5-XL & 3 & 2e-5 & 32 & ~8 & 3 & 1e-4 & 64 & 17 \\
   Llama 2 & 3  & 2e-5 & 32 & ~20 & 3 & 1e-4 & 64 & 21 \\ \midrule
   
    & \multicolumn{8}{c}{\textbf{Finetuning on downstream tasks (Flan-T5-XL)}}  \\\midrule
   ChartQA & 10 & 1e-4 & 128 & ~3 & 10 & 1e-4 & 128 & 7 \\ 
   OpenCQA & 10 & 1e-4 & 128 & ~1.5 & 10 & 1e-4 & 128 & 3 \\
   Chart-to-text Pew & 10 & 1e-4 & 128 & ~2 & 10 & 1e-4 & 128 & 3 \\ 
   Chart-to-text Statista & 10 & 1e-4 & 128 & ~4 & 10 & 1e-4  & 128 & 8 \\ 
   ChartFC & 10 & 1e-4 & 128 & ~1 & 10 & 1e-4 & 128 & 2 \\ \midrule
   
    & \multicolumn{8}{c}{\textbf{Finetuning on downstream tasks (Llama 2)}}  \\\midrule
   ChartQA & 10 & 2e-5 & 32 & ~6 & 10 & 1e-4 & 64 & 8 \\ 
   OpenCQA & 10 & 2e-5 & 32 & ~2 & 10 & 1e-4 & 64 & 4 \\
   Chart-to-text Pew & 10 & 2e-5 & 32 & ~2 & 10 & 1e-4 & 64 & 4 \\ 
   Chart-to-text Statista & 10 & 2e-5 & 32 & ~6 & 10 & 1e-4  & 64 & 9 \\ 
   ChartFC & 10 & 2e-5 & 32 & ~3 &10 & 1e-4 & 64 & 3 \\
    \bottomrule

 \end{tabular}}
 \label{tab:hyperparameters}
\end{table*}

\subsection{Instruction Dataset Analysis}
\label{appendix:wccorpus}
\begin{figure}[t!]
    \centering
    \includegraphics[width=1\textwidth]{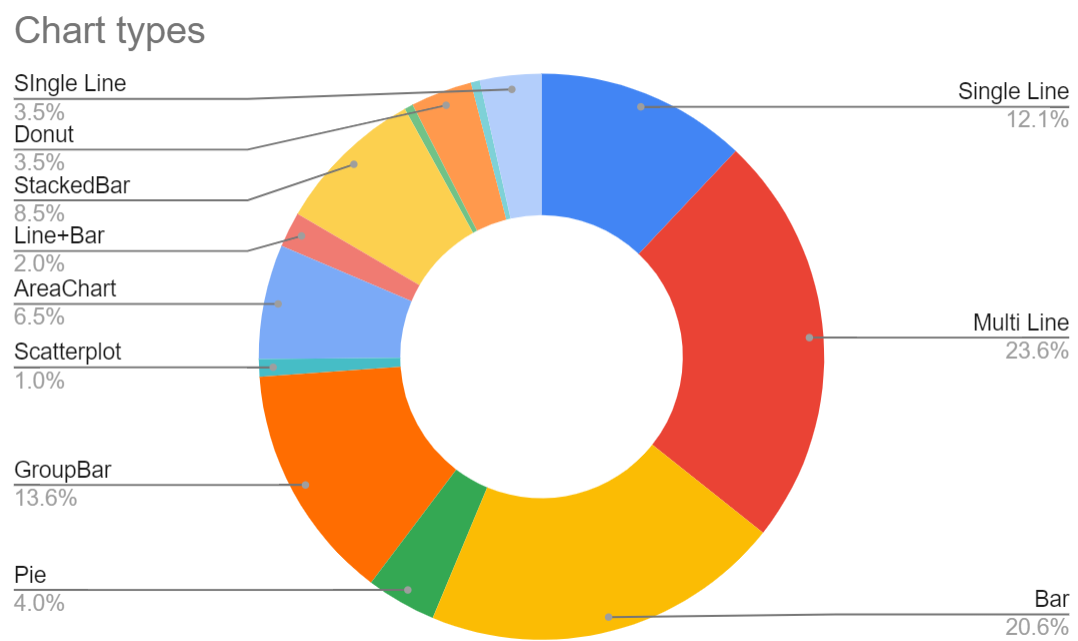}
    \caption{Chart types in WebChart Corpus.}
    \label{fig:chart_types}
\end{figure}

\textbf{WebCharts corpus:} Figure \ref{fig:chart_types} shows the chart type statistics for WebCharts corpus. Since we do not have access to the chart types in this corpus, we manually tagged random  200 images from it to estimate the chart type distribution.

\begin{table}[t]
\setlength\extrarowheight{1pt}
\centering

\scalebox{0.74}{%
\begin{tabular}{l|c}
\toprule
\textbf{Task Group} & \textbf{\#Examples} \\
\midrule
Pattern and Outlier Detection & 16,977 \\
Statistical Analysis & 9,148 \\
Extremum Identification & 6,545 \\
Data Correlations & 6,142 \\
General Comparison & 4,709 \\
Relative Change Calculation & 4,108 \\
Time Series and Future Value Forecasting & 1,944 \\
Data Point Identification & 1,670 \\
Performance and Result Analysis & 1,396 \\
Data Categorization & 533 \\
Distribution analysis & 280 \\
\bottomrule
\end{tabular}%
}

\caption{\small{Number of generated examples for various groups of new tasks created by GPT-4.}}
\label{tab:NovelCounts}
\end{table}

\subsection{Modeling}
\label{appendix:model}
Figure \ref{fig:new_pipeline} shows the pipeline system architecture we use for our models.

\subsection{Ablation Studies}
\label{app:ablations}

To understand the impact of the different instruction tuning tasks on the performance of our model, we conducted ablation studies on the ChartQA dataset using our top performing model, ChartInstruct-Flan-T5-XL (Pipeline System). These ablation experiments involved the removal of one task at a time, except for reasoning tasks, which were grouped and removed together. Due to computational constraints, we finetuned our model on the instruction tuning data for only 1 epoch only, as opposed to the 3 epochs used in the primary experiment. All other hyperparameters remained consistent with those detailed in our main experiments, as outlined in Table \ref{tab:hyperparameters}, including the fine-tuning experiments on the downstream tasks. 

As depicted in Table \ref{tab:ablations}, removing tasks like Chart Summarization and Open-ended Question Answering had a negligible effect on the performance on ChartQA. However, a more significant performance decline was observed upon the exclusion of the fact-checking task, which is important for enhancing the model's data retrieving and reasoning capabilities. This decline was further amplified when reasoning-associated tasks (CoT and Coding) were removed, underscoring their critical role in improving the numerical reasoning capabilities of our model.

\begin{table}[t]
 \centering
 \scalebox{0.73}{\begin{tabular}{l|c}
  
  \toprule
  
   & ChartQA \\ 

   Model & $(RA)$ \\
                                            
    \midrule

  ChartInstruct-Flan-T5-XL & 70.08 \\ 
                                  
  No Open-ended Question Answering  & 69.68 \\ 

  No Chart Summarization & 69.76 \\ 

  No Fact Checking & 66.28 \\

  No Novel Tasks & 69.20 \\ 

  No CoT Reasoning/Coding & 63.36 \\   
 \bottomrule
 
 \end{tabular}}
 \caption{
  \model\ ablations on ChartQA benchmark.
 }
 \label{tab:ablations}
\end{table}

\begin{figure}[t!]
     \centering
        % \vspace{mm}
        \includegraphics[width=0.75\textwidth]{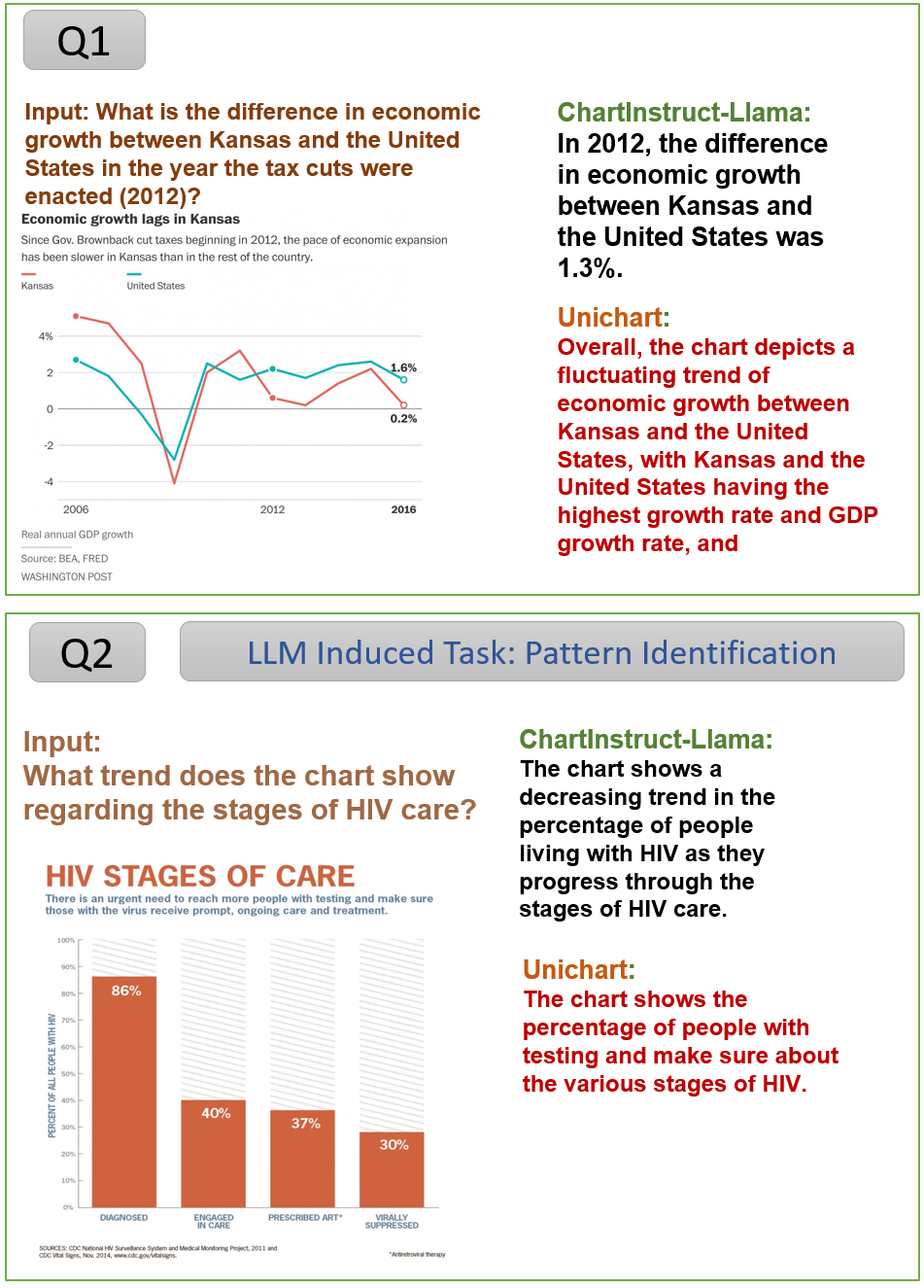}
        \vspace{-1mm}
         \caption{\small{Comparison of \model-Llama and Unichart over two WebChart novel task samples } 
         }\label{fig:modelcomp}
\end{figure}

\subsection{Human Evaluation Study}
Figure \ref{fig:modelcomp} Shows a comparison between \model-Llama and UniChart.

\subsection{Error Analysis}
\begin{figure*}[t!]
     \centering
        % \vspace{mm}
        \scalebox{0.95}{\includegraphics[width=1\textwidth]{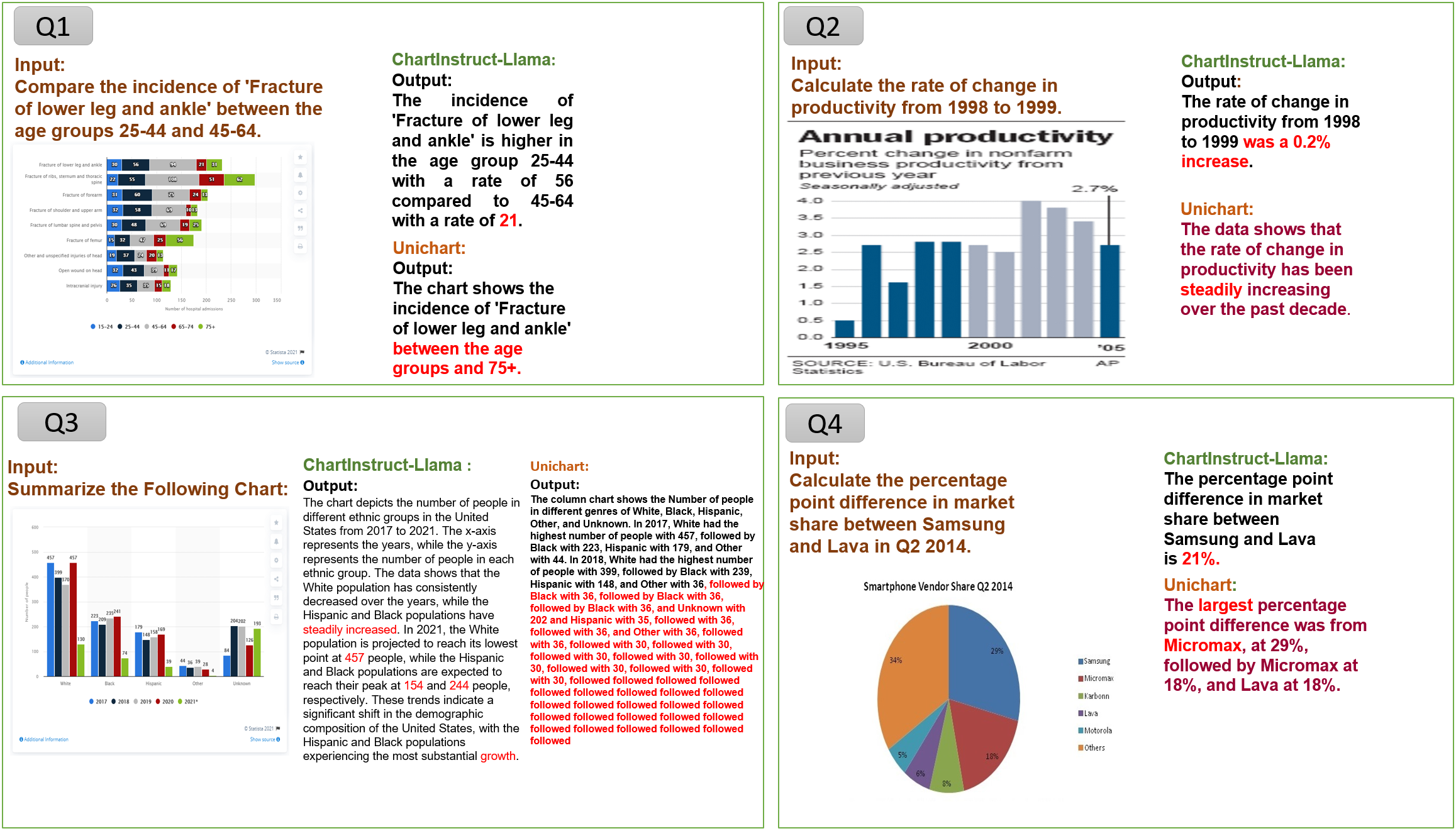}}
        \vspace{-3mm}
         \caption{\small{
         Some samples that our \model-Llama and Unichart had errors in their outputs. The errors are shown in \red{red}. Unichart often does not follow the queries and occasionally repeats tokens. On the other hand, \model-Llama tends to adhere better to the instructions but occasionally encounters factual and other errors.     
         }
         }\label{fig:WrongSamples}
\end{figure*}

Figure \ref{fig:WrongSamples} show a few samples which our model faced a challenge generating factual and accurate responses for. In samples Q1 and Q2, our model fails to find the right value for the expected target either by estimation, matching or comparison to other visual elements. In Q3, Although It generates a cohesive summary, it produces some statements that are not true. Q4 shows a numerical error that \model-Llama didn't perform the subtraction operation correctly.

\begin{figure}[htp]
    \centering
    \includegraphics[width=1\textwidth]{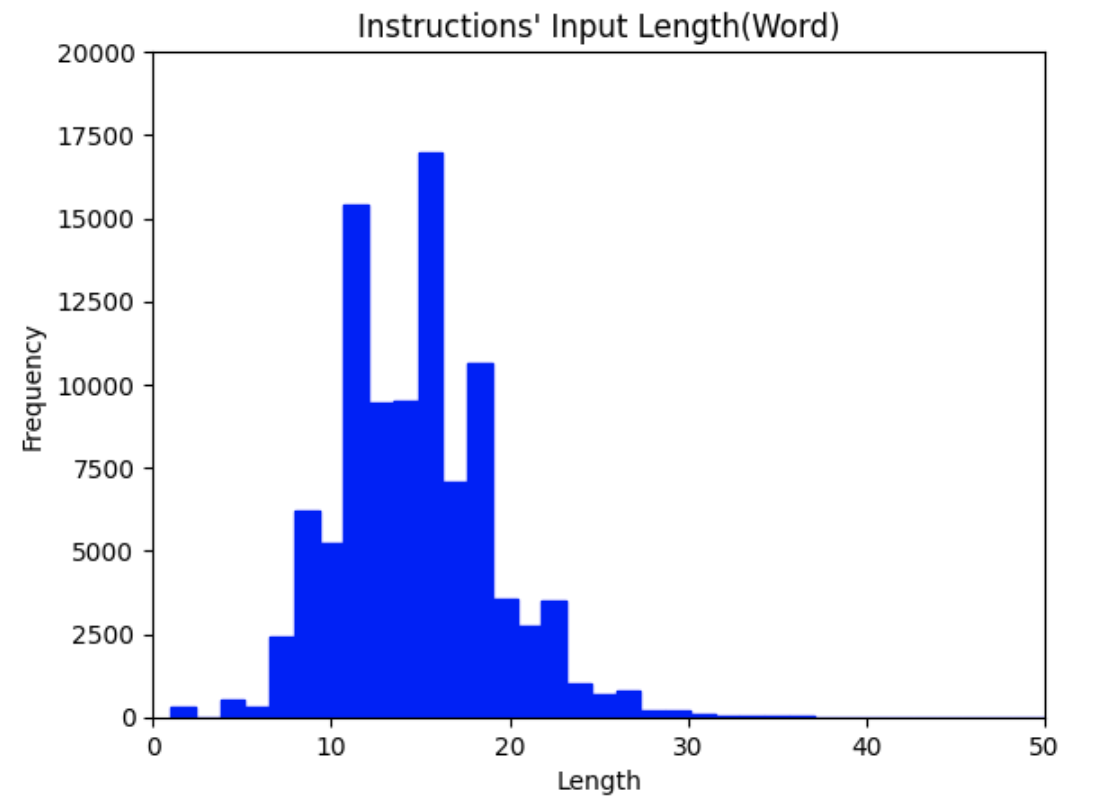}
    \caption{A histrogram of Instruction's Input Length distributions}
    \label{fig:inplen}
\end{figure}

\begin{figure}[htp]
    \centering
    \includegraphics[width=1\textwidth]{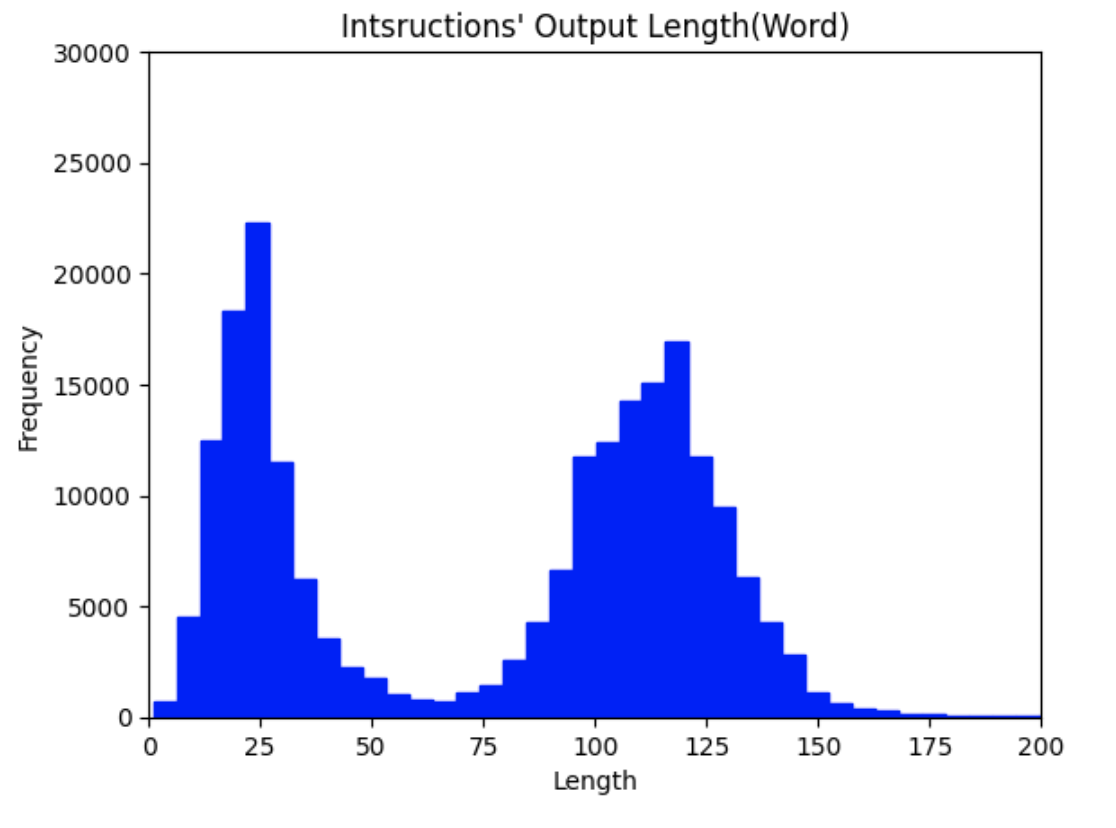}
    \caption{A histrogram of Instruction's Output Length distributions}
    \label{fig:outlen}
\end{figure}

\subsection{Sample Outputs from ChartInstruct}
\begin{figure*}[t!]
    \centering
    \scalebox{0.95}{\includegraphics[width=1\textwidth]{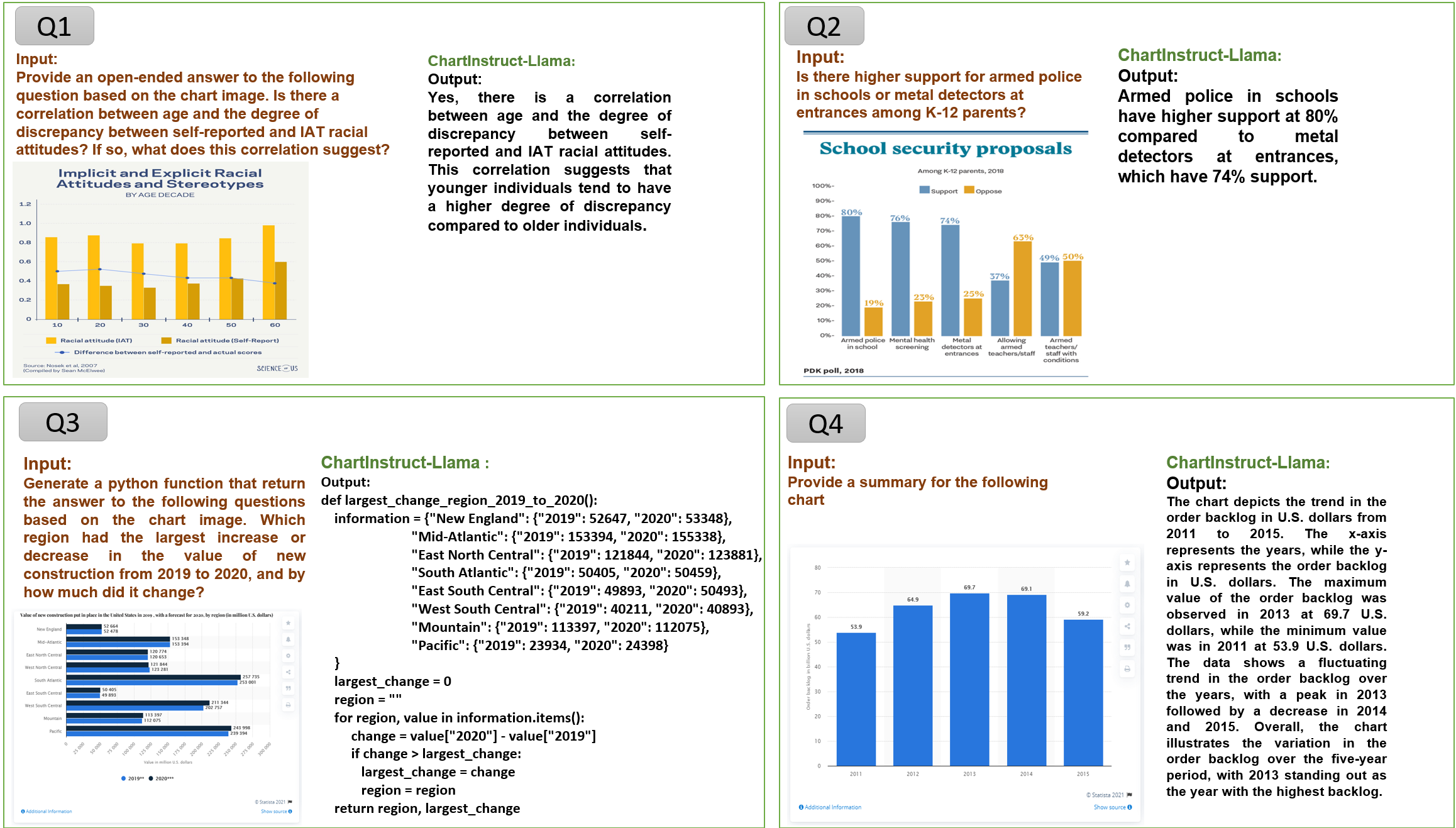}}
    \caption{Sample outputs generated by \model-Llama over various tasks such as LLM induced Data Correlation, Open-EndedCQA, Coding, and Chart Summarization}
    \label{fig:goods}
\end{figure*}

In Figure\ref{fig:goods}, we provide some sample outputs on various tasks.

\begin{figure*}[t!]
     \centering
        % \vspace{mm}
        \scalebox{0.95}{\includegraphics[width=1\textwidth]{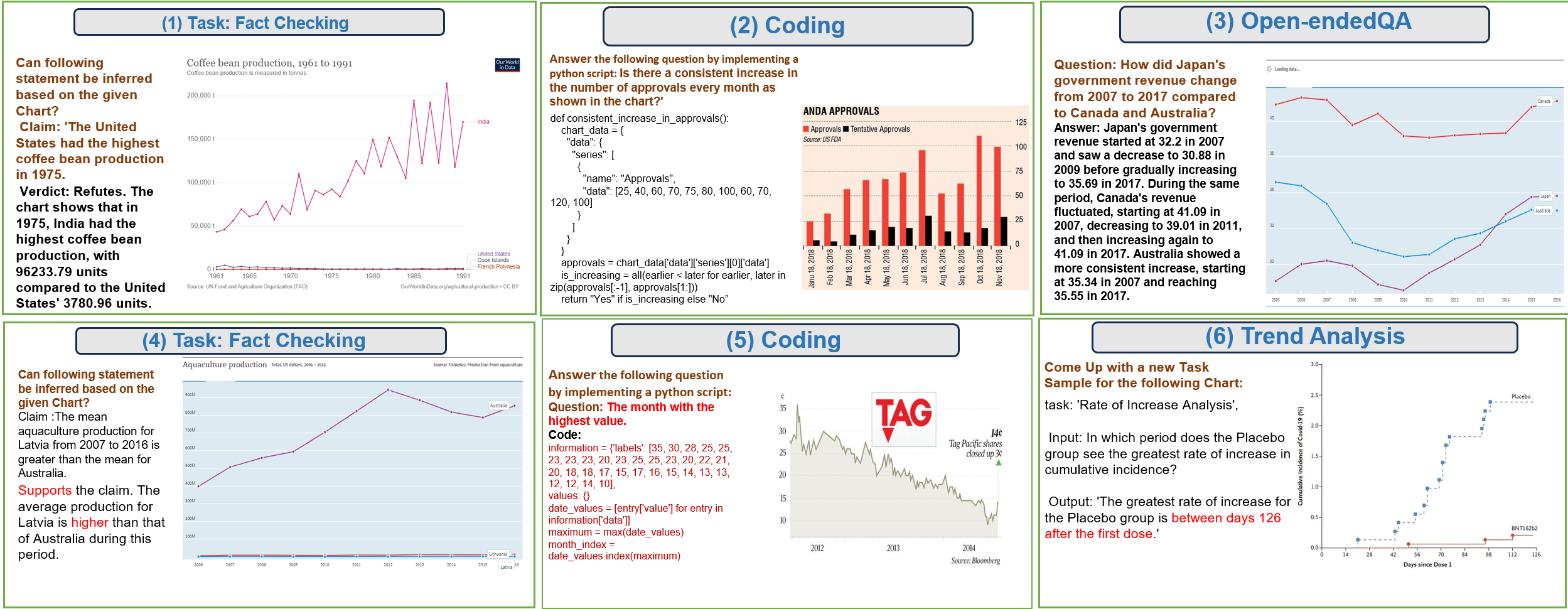}}
        \vspace{-3mm}
         \caption{\small{
         Few generated instruction samples. The Top row shows correct generated instructions. The bottom row shows some of the instructions involve errors: (4) generated wrong answer, (5) not following the description - wrong answer, (6) wrong answer. The errors are shown in \red{red}. 
         }
         }\label{fig:few-generated-samples}
\end{figure*}

\subsection{Generated Data Samples}
\label{appendix:generatedsamples} 

The input prompt for each specific task is provided through Tables \ref{tab:FCsample}-\ref{tab:Novelsample}.

\begin{table*}[t]
\centering
\caption{An example of prompt for Fact Checking Instruction Generation using GPT-3.5-turbo and an extracted sample}
\label{tab:FCsample}
\scalebox{0.7}{
\begin{tabular}{p{0.7\linewidth}p{0.4\linewidth}}
\Xhline{4\arrayrulewidth}
\multicolumn{1}{l}{\textbf{Fact Checking Prompt for GPT-3.5}} & \\ \hline
\multicolumn{1}{p{0.9\linewidth}}{{\small \textbf{Prompt:} \newline

Analyze a chart and generate pairs of claims and verdicts about its data.
Half of the claims should be supported by the chart's data, while the other 
half are refuted. Avoid using terms like 'rows', 'columns', or 'elements' from 
the data table; refer to 'chart' or 'chart image' instead. Each response must 
state whether it 'supports' or 'refutes' the question, followed by a brief 
explanation. The claims should cover comparisons of values or trends, basic 
statistical values (maximum, minimum, mean, median, mode) without using
exact numbers from the chart. Ensure a diverse range of claims addressing various visual aspects of the chart, resulting in 5-6 turns of C\&V.\
Put \~{} before starting each claim, and \^{} before each answer.\newline
\blue{Title:} \newline Angola: Population from 2009 to 2019 by gender(in million) \newline
\blue{Table:} \newline Characteristic, 	 Female, 	 Male \newline
2019,	16.08,	15.74\newline
2018,	15.57,	15.24\newline
2017,	15.07,	14.75\newline
2016,	14.58,	14.27\newline
2015,	14.09,	13.79\newline
2014,	13.62,	13.32\newline
2013,	13.15,	12.86\newline
2012,	12.69,	12.41\newline
2011,	12.25,	11.97\newline
2010,	11.81,	11.55\newline
2009,	11.38,	11.14\newline
\blue{Extracted Instruction:}\newline
Claim: \newline
In 2019, there was a significant decrease in the total population of males compared to 2018.\newline
Verdict: \newline
Refutes. The chart indicates a slight increase in the population of males from 2018 to 2019, as the value for males in 2019 (15.74 million) is higher than that
in 2018 (15.24 million), contradicting the claim of a significant decrease.
} }& 
\raisebox{-\height}{\includegraphics[width=0.4\textwidth]{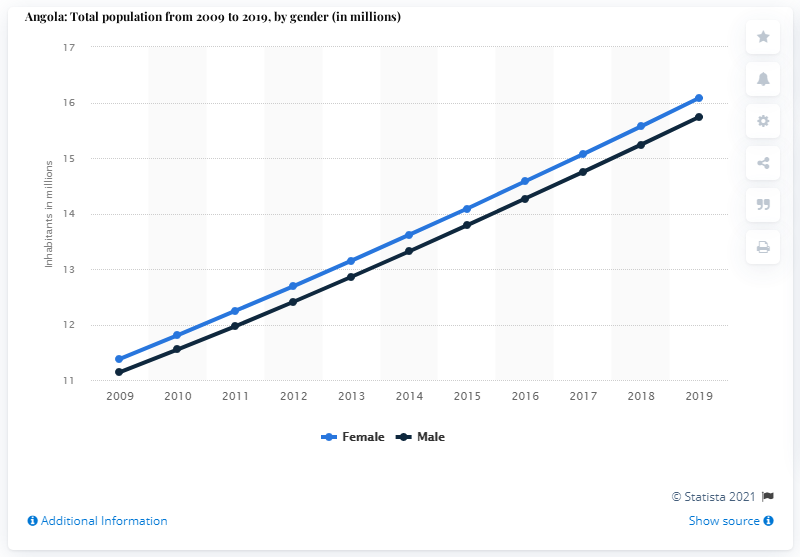}} \\
\multicolumn{2}{l}{} \\
\multicolumn{2}{p{\linewidth}}{} \\
\multicolumn{2}{p{\linewidth}}{} \\ \hline
\end{tabular}
}
\end{table*}

\begin{table*}[t]
\centering
\caption{An example of prompt for Summarization Instruction Generation using GPT-3.5-turbo and an extracted sample}
\label{tab:Summsample}
\scalebox{0.65}{
\begin{tabular}{p{0.7\linewidth}p{0.4\linewidth}}
\Xhline{4\arrayrulewidth}
\multicolumn{1}{l}{\textbf{Summarization Prompt for GPT-3.5}} & \\ \hline
\multicolumn{1}{p{0.9\linewidth}}{{\small \textbf{Prompt:} \newline

Summarize a chart based on a provided data table, highlighting key aspect such as data background, maximum and minimum values, value comparisons, and 
trends across categories. The summary should not directly mention the input
table, instead referring to the 'chart' or 'chart elements'. You'll also be given the title of the chart to get a context on what is shown in the chart.
Do not use the title as it is like "the chart is titles TITLE", always try to 
infersomething new based on it. Include a brief explanation of the elements represented on the x-axis and y-axis. The entire summary should be cohesive, in
paragraph form without bullet points, and not exceed 150 words. Conclude with a 
final sentence that encapsulates the most significant findings from the chart.\newline
\blue{Title:} \newline Title: Annual Change in Oil Consumption \newline
\blue{Table:} \newline
Characteristic 	 Romania 	 Uzbekistan 	 Slovenia 	 Australia 	 Canada \newline
1966	5.31,	  -,  	  -,  	34.29,	35.92 \newline
1970	11.43,	  -,  	  -,  	28.99,	54.29 \newline
1980	-10.61,	  -,  	  -,  	-13.87,	-15.1 \newline
1990	19.19,	-36.32,	  -,  	7.35,	-12.24 \newline
2000	7.35,	1.64,	0.01,	-0.81,	1.64 \newline
2010	4.5,	-5.71,	1.64,	34.29,	31.84 \newline
\blue{Extracted Instruction:}\newline
Summary:\newline
The chart depicts the annual change in oil consumption across five countries 
over several decades. On the x-axis, time spans from 1966 to 2010, while the 
y-axis represents the percentage change in oil consumption. Each line on the 
chart represents a different country: Romania, Uzbekistan, Slovenia,
Australia, and Canada. Maximum values are observed in Australia and Canada in 
various years, while Romania shows relatively stable fluctuations. Uzbekistan
experiences significant fluctuations, including a notable decrease in 1990. 
Slovenia and Canada display a generally increasing trend. Overall, Australia 
and Canada exhibit the highest peaks and most consistent growth in oil 
consumption over the period analyzed.

} }& 
\raisebox{-\height}{\includegraphics[width=0.4\textwidth]{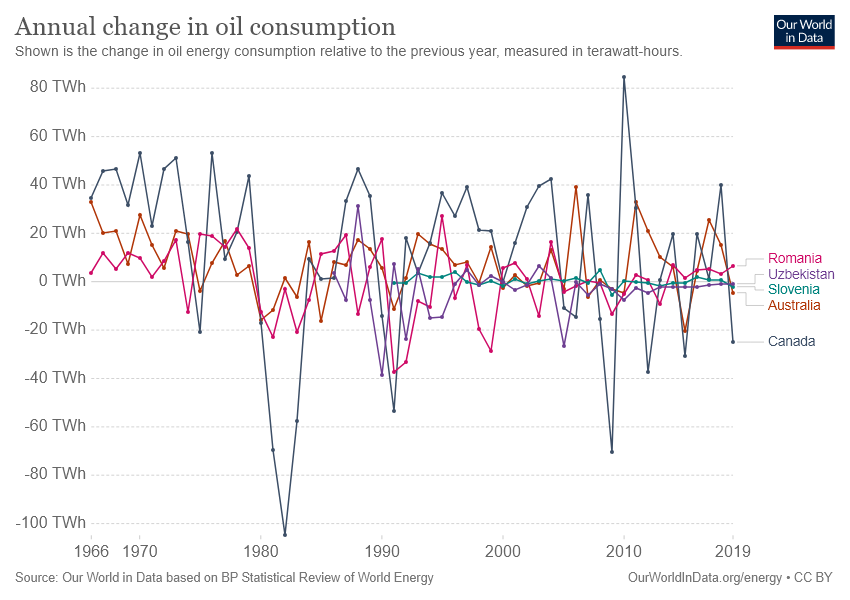}} \\
\multicolumn{2}{l}{} \\
\multicolumn{2}{p{\linewidth}}{} \\
\multicolumn{2}{p{\linewidth}}{} \\ \hline
\end{tabular}
}
\end{table*}

\begin{table*}[t]
\centering
\caption{An example of  prompt for OCQA Instruction Generation using GPT-3.5-turbo and an extracted sample}
\label{tab:OCQAsample}
\scalebox{0.7}{
\begin{tabular}{p{0.7\linewidth}p{0.4\linewidth}}
\Xhline{4\arrayrulewidth}
\multicolumn{1}{l}{\textbf{OCQA Prompt for GPT-3.5}} & \\ \hline
\multicolumn{1}{p{0.9\linewidth}}{{\small \textbf{Prompt:} \newline

Generate question-answer pairs based on a data table representing a chart.
You will also be given the title to have additional context. The questions 
should vary in complexity, ranging from simple 'what', 'which', 'when' 
questions to more challenging 'how' and 'why' questions. Refer to 'chart' or
'chart elements' instead of the data table. Ensure the answers are cohesive
and fluent, using diverse vocabulary, with each answer spanning 4 to 5 
sentences. The questions should encompass different aspects related to a chart figure. Each turn of conversation should consist of 5 or 6 question-answer 
pairs, covering a range of insights, explanations, and complexities as mentioned. 
Try to avoid generic questions such as what is the title or subtitle of the 
charts. Having questions with more challenging targets is preferable. There's no specific order for the difficulty of the questions. Put \~{} before starting each question, and \^{} before each answer.\newline
\blue{Title: }\newline
How often do you read magazinzs(by age) \newline
\blue{Table:}\newline
Characteristic 	 Never 	 Rarely 	 Occasionally 	 Often \newline
 18-29 	9\%	30\%	34\%	27\%\newline
 30-39 	6\%	14\%	39\%	41\%\newline
 40-49 	15\%	29\%	33\%	23\%\newline
 50+ 	12\%	24\%	44\%	20\%\newline
What is the trend in magazine reading frequency among different age groups?\newline
Magazine reading frequency varies significantly across age groups. Young \newline
adults aged 18-29 predominantly fall into the "Occasionally" category, with \newline34\% indicating they read magazines occasionally. In contrast, those \newline aged 30-39 and 40-49 are more likely to read magazines "Often," with 39\%\newline
and 33\% respectively. However, among individuals aged 50 and above, the \newline frequency of magazine reading declines, with only 20\% indicating they \newline read often.

} }& 
\raisebox{-\height}{\includegraphics[width=0.4\textwidth]{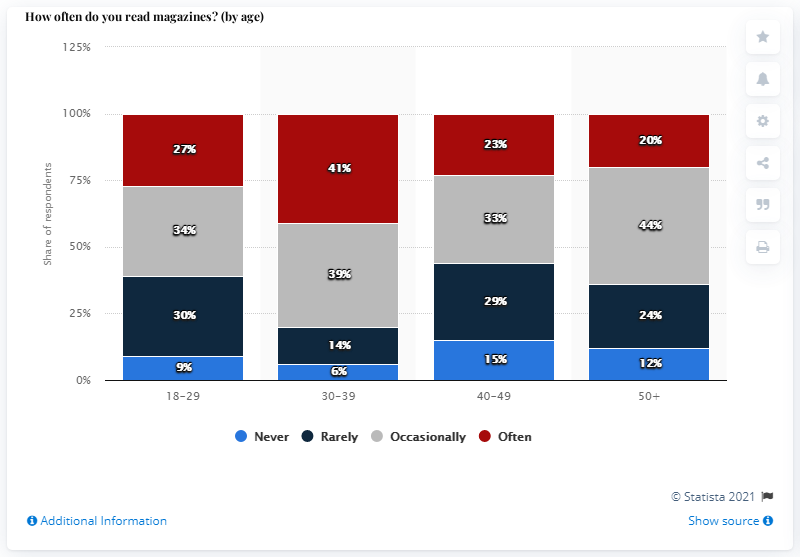}} \\
\multicolumn{2}{l}{} \\
\multicolumn{2}{p{\linewidth}}{} \\
\multicolumn{2}{p{\linewidth}}{} \\ \hline
\end{tabular}
}
\end{table*}

\begin{table*}[t]
\centering
\caption{An example of  Variable-Independent prompt for COT instruction Generation for GPT-4-preview and an extracted sample}
\label{tab:COTVIsample}
\scalebox{0.7}{
\begin{tabular}{p{0.7\linewidth}p{0.4\linewidth}}
\Xhline{4\arrayrulewidth}
\multicolumn{1}{l}{\textbf{COT Prompt for GPT-4}} & \\ \hline
\multicolumn{1}{p{0.9\linewidth}}{{\small \textbf{Prompt:} \newline

You’ll be given a data table which is an underlying data table for a chart. 
Although you can’t see the chart, you should mention any references that come
up to the chart instead of the table. Your responses must be limited to
question-and-answer pairs related to the data, strictly avoiding any 
conversational language for explanation or fillers. Questions must be at 
most two-part answer. You have to come up with six pairs of questions and 
answers that ask for a chart value, or comparisons not asking for a number,
and a general trend. There must be at least two retrieval questions and one
general trend. To come up with your answer, you have to break down your
solution into several basic steps. In each step, you explain what are you
going to do, and if that is a mathematical operation, you have to mention the
formula and fill it with the respective numbers. Avoid using = operator in your
calculations. If you are comparing two or several numbers retrieved from the 
table, you must mention them. Consider the following template: " The Answer is <ANSWER>." to wrap up your generated steps, replace <ANSWER> in the mentioned 
template with your final answer word in the most minimalistic way without ANY 
explanations or assignment. Please return the list of questions and responses 
in json format. I should be able to parse it. Each dictionary must contain 3 field 'question', 'steps', 'answer' filled with the requested formation.\newline
\blue{Title:} \newline Average price od diesel fuel in France between January 2023 and December 2018(in euroes per liter) \newline
\blue{Table:} \newline
Characteristic 	2013	2014	2015	2016	2017	2018\newline
 Dec 	1.39	1.16	1.11	1.18	1.28	1.42 \newline
 Nov 	1.39	1.31	1.17	1.24	1.27	1.47 \newline
 Oct 	1.41	1.36	1.22	1.25	1.24	1.53 \newline
 Sep 	1.43	1.36	1.18	1.19	1.22	1.48 \newline
 Aug 	1.41	1.38	1.21	1.16	1.2	1.45 \newline
 Jul 	1.43	1.39	1.27	1.21	1.17	1.45 \newline
 Jun 	1.41	1.39	1.28	1.21	1.19	1.46 \newline
 May 	1.41	1.38	1.3	1.07	1.2	1.44 \newline
 Apr 	1.46	1.38	1.27	1.28	1.25	1.4 \newline
 Mar 	1.46	1.41	1.39	1.13	1.26	1.37 \newline
 Feb 	1.49	1.41	1.21	1.08	1.26	1.39 \newline
 Jan 	1.46	1.4	1.17	1.1	1.27	1.4 \newline
 \blue{Extracted output: }\newline
 \{
    "question": "What is the trend in diesel fuel prices from January to December in 2018?", \newline 
    "steps": "To determine the trend, observe the prices from January to December in 2018. Start with January (1.4) and end with December (1.42), 
    noting any significant increases or decreases throughout the year.", \newline
    "answer": "The Answer is increase."
    \}

} }& 
\raisebox{-\height}{\includegraphics[width=0.4\textwidth]{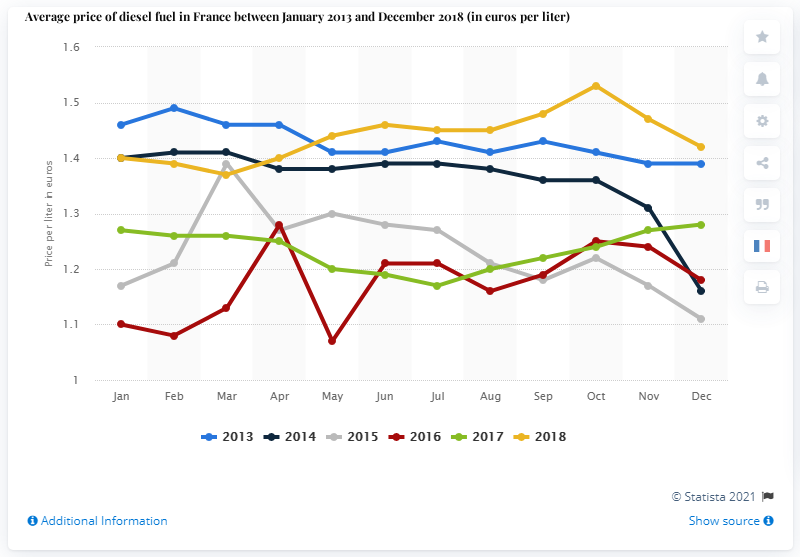}} \\
\multicolumn{2}{l}{} \\
\multicolumn{2}{p{\linewidth}}{} \\
\multicolumn{2}{p{\linewidth}}{} \\ \hline
\end{tabular}
}
\end{table*}

\begin{table*}[t]
\centering
\caption{An example of Variable Dependent prompt for Chain of Thought Instruction Generation using GPT-4 and an extracted sample}
\label{tab:VOTVDsample}
\scalebox{0.7}{
\begin{tabular}{p{0.7\linewidth}p{0.4\linewidth}}
\Xhline{4\arrayrulewidth}
\multicolumn{1}{l}{\textbf{COT Prompt for GPT-4-preview}} & \\ \hline
\multicolumn{1}{p{0.9\linewidth}}{{\small \textbf{Prompt:} \newline

First of all, know that you must act like you can't do basic match even
addition or subtraction. However, you might feel a need for calculation 
in your incoming tasks. In that case, you can tell me to calculate by calling 
Calculator(X = OPERATION). <OPERATION> can be executable python code and X is
a variable. Use the defined variables in the next sentences when needed like 
a python code.  Do not define the variables in the text. Use this format "DEFINE(variable\_name=X)" where X is the value of the variable. You’ll be  given a datatable in json string format which is an underlying datatable for
a chart. Although you can’t see the chart, you should mention any references 
that come up to the chart instead of the table. Your responses must be limited 
to question and answer pairs related to the data, strictly avoiding any 
conversational language or fillers. You have to come up with six pairs of 
question and answers that ask for values that require calculations such as 
differences, totals, and statistical measures such as median, mode, and mean.
To come up with your answer, you have to break down your solution into several
basic steps. In each step, you explain what are you going to do, and if that is 
a mathematical operation, you have to mention the formula and fill it with the 
respected numbers. If you want to list some numbers, do it when there are less 
than 8 of them. Since you are unable to do calculations, you may return one of 
the defined variables from your previous steps as your final answer. Consider 
the following template: " The Answer is <ANSWER>." to wrap up your generated 
steps, replace <ANSWER> in the mentioned template with your final answer word 
without ANY explanations or assignment. Please return the list of questions 
and responses in a json format. I should be able to parse it. Each dictionary 
must contain 3 field 'question', 'steps', 'answer' filled with the requested 
formation. The answer section must only involve a python variable previously 
defined. No numbers should appear in this section.\newline

\blue{Title: } \newline
 Driver Satisfaction with Uber in United States from 2017 to 2019 \newline
 \blue{Table: }\newline
Characteristic 	2017	2018	2019 \newline
 Strongly disagree 	10.80\%	8.20\%	13.90\% \newline
 Somewhat disagree 	25.40\%	17.30\%	22.50\% \newline
 Neither agree nor disagree 	14.40\%	16.30\%	19.10\% \newline
 Somewhat agree 	39.70\%	43.40\%	34.10\% \newline
 Strongly agree 	9.70\%	14.80\%	10.50\% \newline
 \blue{Extracted Output:} \newline 
  \{
    "question": "What is the increase in the percentage of drivers who
    somewhat agree from 2017 to 2018?",\newline
    "steps":
      "DEFINE(somewhat\_agree\_increase=43.4-39.7)",
      "First, I will subtract the percentage of drivers who somewhat agree in 2017, which is 39.7\%, from the percentage of drivers who somewhat agree in 2018, which is 43.4\%.", \newline
    "answer": "The Answer is somewhat\_agree\_increase."
  \}

} }& 
\raisebox{-\height}{\includegraphics[width=0.4\textwidth]{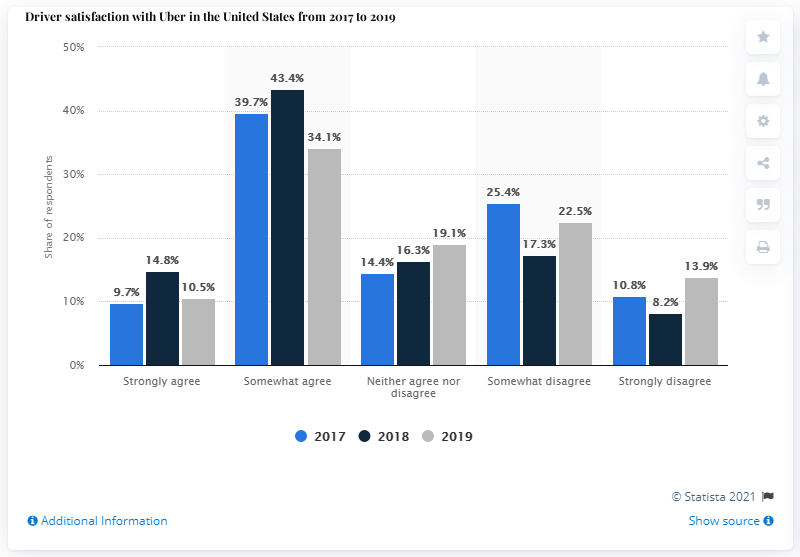}} \\
\multicolumn{2}{l}{} \\
\multicolumn{2}{p{\linewidth}}{} \\
\multicolumn{2}{p{\linewidth}}{} \\ \hline
\end{tabular}
}
\end{table*}

\begin{table*}[t]
\centering
\caption{An example of our coding prompt for GPT-3.5 and an extracted sample}
\label{tab:code3sample}
\scalebox{0.65}{
\begin{tabular}{p{0.7\linewidth}p{0.4\linewidth}}
\Xhline{4\arrayrulewidth}
\multicolumn{1}{l}{\textbf{Coding Prompt for GPT-3.5}} & \\ \hline
\multicolumn{1}{p{0.9\linewidth}}{{\small \textbf{Prompt:} You are a programmer expert in python. I want to analyze chart data. However,
I can only provide you the underlying data as json line format representing 
the chart instead of its actual image. The underlying data involves different elements. I want you to come up with pairs of insightful questions, and a python function that answers the insightful question. Here are the requirements
for the questions and answers:\newline
- Do not refer to the table in the question. Mention " from the chart" if needed. \newline
- Your output is limited to several questions followed by a Python script, strictly avoid providing any explanation before or after questions and answers such as stating the question type.\newline
- Use \~{} at the beginning of each question and \^{} after finishing each questions. \newline
- question and answers should appear one by one in the following format: ~Question\^{}, Python Script, ~Question\^{},  Python Script.\newline
- Nothing in English should appear in the output other than the questions.\newline
- Act like you don't have access to the table I have given you. As a result, you must define the given data at the beginning of each function.\newline
- Inside each generated function, Define a dictionary called information that involves the data split such as labels and values.\newline
- The Python script must only be a function and not anything else such as writing a body for the code and calling the function.\newline
- The function should be independent and not rely on any sources out of that function.\newline
- The generated script must be executable by Python, so avoid writing anything such as ... or // in between.\newline
- The functions must not receive any input arguments. \newline
- Everything that is needed must be defined in the function.\newline
- Variable independent are defined as follows: Questions that want to return a\newline string as response. The funtion must return a string which either "Yes"/"No" based on the conditions, trends, or a string that explains something like which characteristic does this number belong to?\newline
-  Variable dependent questions are defined as follows: Questions that ask for either a characteristic which is not known and must be found, or a statistical metric and measures that must be calculated. The output would be either a calculated number or a variable.\newline
- Retrieval Questions: Questions that ask for characteristics or values represented in the chart.\newline
- Hard questions: Questions that involve applying multiple filters and calculating advanced statistical metrics.\newline
Come up with 8 question for each table in the following order, two variable independent questions, and a mix of three retrieval and variable dependent questions and three hard questions.\newline
\blue{Title:} \newline Percentage of Population in England who gave to charity from 2013/14 to 2019/20 by age group\newline
\blue{Table:}\newline
Characteristic 	 16 to 24 	 25 to 34 	 35 to 49 	 50 to 64 	 65 to 74 	 75 and over \newline
 2013/14 	74\%	77\%	82\%	84\%	89\%	89\%\newline
 2014/15 	63\%	70\%	79\%	82\%	85\%	89\%\newline
 2015/16 	60\%	70\%	76\%	83\%	85\%	83\%\newline
 2016/17 	57\%	68\%	78\%	80\%	85\%	85\%\newline
 2017/18 	57\%	69\%	75\%	80\%	85\%	83\%\newline
 2018/19 	59\%	69\%	76\%	79\%	82\%	83\%\newline
 2019/20 	55\%	66\%	77\%	79\%	84\%	85\%\newline
 
Extracted Output:\newline
 Question: What is the standard deviation of the percentage of population giving to charity for the age group 35 to 49 across all years?\newline
 Answer:\newline
 def std\_deviation\_age\_group\_3549():\newline
    data = [
        \{"year": "2013/14", "16 to 24": 74, "25 to 34": 77, "35 to 49": 82, "50 to 64": 84, "65 to 74": 89, "75 and over": 89\},\newline
        \{"year": "2014/15", "16 to 24": 63, "25 to 34": 70, "35 to 49": 79, "50 to 64": 82, "65 to 74": 85, "75 and over": 89\},\newline
        \{"year": "2015/16", "16 to 24": 60, "25 to 34": 70, "35 to 49": 76, "50 to 64": 83, "65 to 74": 85, "75 and over": 83\},\newline
        \{"year": "2016/17", "16 to 24": 57, "25 to 34": 68, "35 to 49": 78, "50 to 64": 80, "65 to 74": 85, "75 and over": 85\},\newline
        \{"year": "2017/18", "16 to 24": 57, "25 to 34": 69, "35 to 49": 75, "50 to 64": 80, "65 to 74": 85, "75 and over": 83\},\newline
        \{"year": "2018/19", "16 to 24": 59, "25 to 34": 69, "35 to 49": 76, "50 to 64": 79, "65 to 74": 82, "75 and over": 83\},\newline
        \{"year": "2019/20", "16 to 24": 55, "25 to 34": 66, "35 to 49": 77, "50 to 64": 79, "65 to 74": 84, "75 and over": 85\}\newline
    ]\newline
    information = \{"years": [d["year"] for d in data], "age\_groups": ["16 to 24", "25 to 34", "35 to 49", "50 to 64", "65 to 74", "75 and over"]\}\newline
    percentages = [d["35 to 49"] for d in data]\newline
    mean = sum(percentages) / len(percentages)\newline
    variance = sum((x - mean) ** 2 for x in percentages) / len(percentages)\newline
    std\_dev = variance ** 0.5\newline
    return std\_dev
} }& 
\raisebox{-\height}{\includegraphics[width=0.4\textwidth]{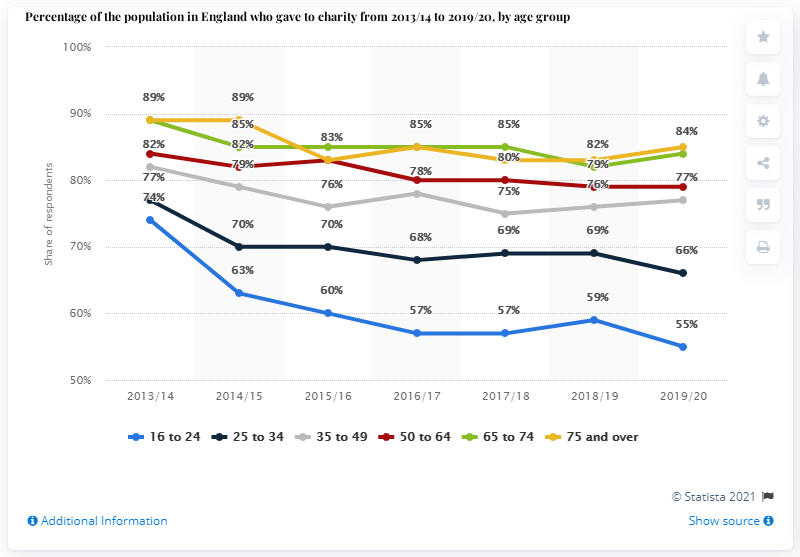}} \\
\multicolumn{2}{l}{} \\
\multicolumn{2}{p{\linewidth}}{} \\
\multicolumn{2}{p{\linewidth}}{} \\ \hline
\end{tabular}
}
\end{table*}

\begin{table*}[t]
\centering
\caption{An example of  prompt for Novel Instruction Generation for GPT-4-preview and an extracted sample}
\label{tab:Novelsample}
\scalebox{0.65}{
\begin{tabular}{p{0.7\linewidth}p{0.4\linewidth}}
\Xhline{4\arrayrulewidth}
\multicolumn{1}{l}{\textbf{Novel Prompt for GPT-4}} & \\ \hline
\multicolumn{1}{p{0.9\linewidth}}{{\small \textbf{Prompt:} \newline

Generate different instruction-tuning tasks for an LLM that we are trying to tune for 
Chart Understanding. Your response should be in a json format where each example has
three fields: task type, input: which only asks a question related to the task type and
the given data, and expected output: which is the answer to the input question based on the input information. Use the following chart data to generate 10 unique tasks that do not overlap with two specific tasks: Summarizing the Chart, and single point information retrieval.\newline
\blue{Title:} Median Feature Film Budget In the United Kingdom (UK) from 2008 to 2019, by production type(in million GBP)\newline
\blue{Table:}\newline
Production category 	 Inward investment films 	 Domestic UK films 	 Co-productions \newline
2008	17.2	0.2	1.4\newline
2009	18.45	0.2	1.35\newline
2010	13.09	0.13	2.5\newline
2011	17.64	0.18	1.1\newline
2012	3	0.15	1.03\newline
2013	10.84	0.16	1.06\newline
2014	10.32	0.25	1.44\newline
2015	6.07	0.28	0.9\newline
2016	8.23	0.29	2\newline
2017	9.33	0.3	3.51\newline
2018	10.4	0.51	2.57\newline
2019	7.7	0.75	3.22\newline
\blue{Extracted Instruction:}\newline
    "Task: "Range Detection", \newline
  "input":What range of median budgets did domestic UK films have over the period?\newline "expected\_output": "Domestic UK films had median budgets ranging from 0.13 million GBP in 2010 to 0.75 million GBP in 2019."\newline

} }& 
\raisebox{-\height}{\includegraphics[width=0.4\textwidth]{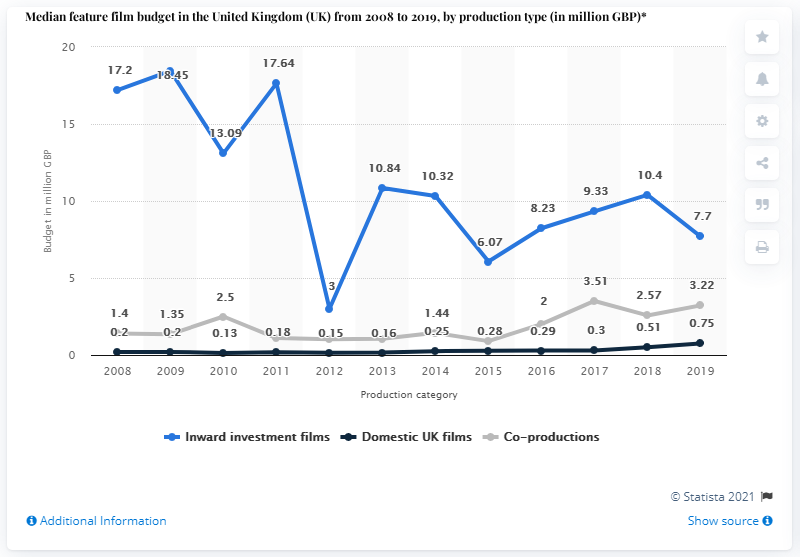}} \\
\multicolumn{2}{l}{} \\
\multicolumn{2}{p{\linewidth}}{} \\
\multicolumn{2}{p{\linewidth}}{} \\ \hline
\end{tabular}
}
\end{table*}

\begin{figure*}[t!]
    \centering
    \includegraphics[width=1\textwidth]{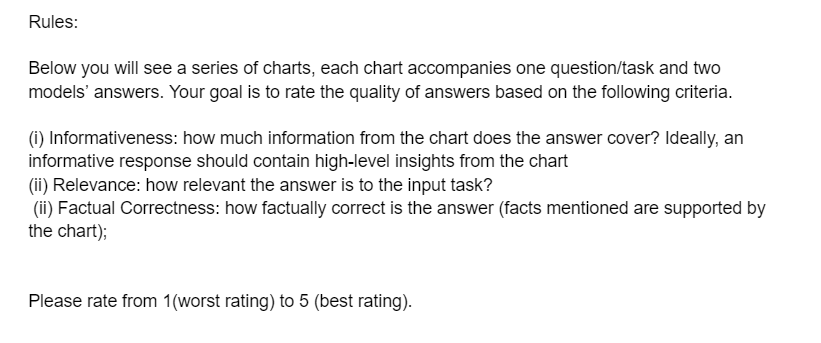}
    \caption{Evaluation Rules of the human study for the three following metrics: informativeness, relevance, and factual correctness.}
    \label{fig:evalRule}
\end{figure*}

\begin{figure*}[t!]
    \centering
    \includegraphics[width=1\textwidth]{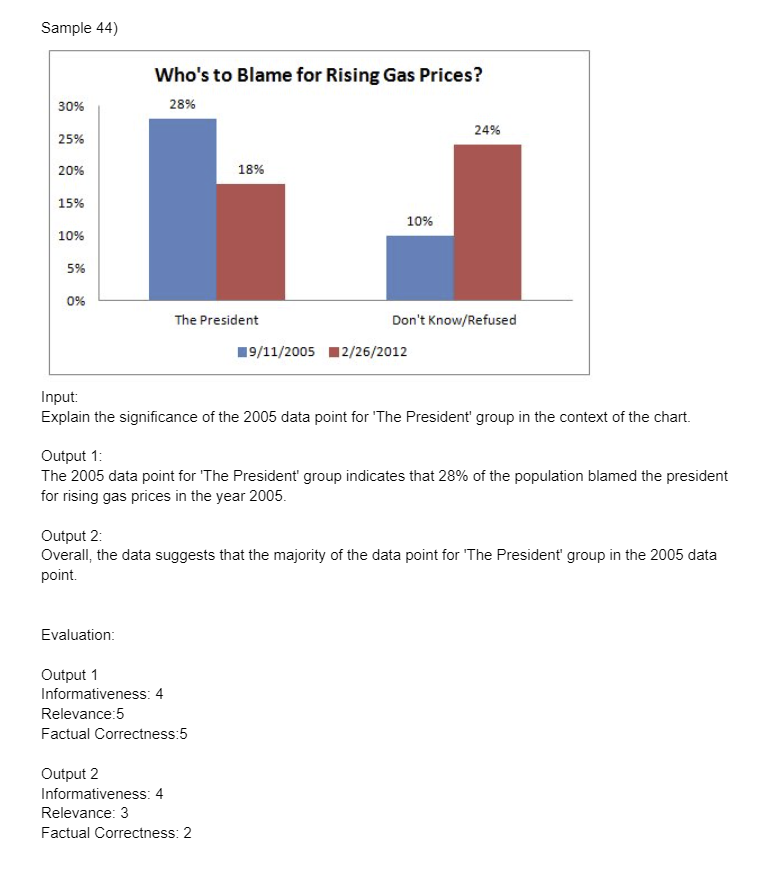}
    \caption{An example of the human study evaluation setup.}
    \label{fig:evalenv}
\end{figure*}

\end{appendices}

\end{document}